\documentclass{article}

\usepackage{microtype}
\usepackage{graphicx}
\usepackage{booktabs} 

\usepackage{hyperref}

\usepackage[accepted]{icml2024}

\usepackage{amsmath}
\usepackage{amssymb}
\usepackage{mathtools}
\usepackage{amsthm}

\usepackage[utf8]{inputenc} 
\usepackage[T1]{fontenc}    
\usepackage{hyperref}       
\usepackage{url}            
\usepackage{booktabs}       
\usepackage{amsfonts}       
\usepackage{nicefrac}       
\usepackage{microtype}      
\usepackage{algorithmic}
\usepackage{algorithm}
\usepackage{tikz}
\usepackage{multirow}
\usepackage{booktabs}
\usepackage{tabularx}
\usepackage{enumitem}
\setlist[itemize]{topsep=0pt, partopsep=0pt, parsep=0pt, itemsep=6pt}
\usepackage{subcaption}

\usepackage{xcolor}

\definecolor{pastel-red}{HTML}{FF9AA2}
\definecolor{pastel-green}{HTML}{B5EAD7}
\definecolor{pastel-blue}{HTML}{A7DBD8}
\definecolor{pastel-teal}{HTML}{9AD9DB}
\definecolor{pastel-yellow}{HTML}{FFFFCC}
\definecolor{pastel-purple}{HTML}{D3C0F9}
\definecolor{pastel-orange}{HTML}{FFC2A2}
\definecolor{pastel-pink}{HTML}{FFB7B2}
\definecolor{pastel-cyan}{HTML}{A1E8E2}
\definecolor{pastel-lavender}{HTML}{E0BBE4}

\usepackage{amsfonts}
\usepackage{color}
\newcommand{\xxcomment}[4]{\textcolor{#1}{[$^{\textsc{#2}}_{\textsc{#3}}$ #4]}}
\newcommand{\ben}[1]{\xxcomment{pastel-blue}{B}{A}{#1}}

\usepackage{url}
\usepackage{verbatim}
\usepackage{xspace}
\usepackage{upgreek}

\usepackage[capitalize,noabbrev]{cleveref}

\theoremstyle{plain}

\theoremstyle{definition}

\theoremstyle{remark}

\usepackage[textsize=tiny]{todonotes}

\usepackage{url}
\usepackage{verbatim}
\usepackage{xspace}
\usepackage{upgreek}
\usepackage{amsmath}

\usepackage[capitalize,noabbrev]{cleveref}
\usepackage{soul}

\usepackage{listings}

 \definecolor{codegreen}{rgb}{0,0.6,0}
\definecolor{codegray}{rgb}{0.5,0.5,0.5}
\definecolor{codepurple}{rgb}{0.58,0,0.82}
\definecolor{backcolour}{rgb}{0.95,0.95,0.92}

 \lstdefinestyle{mystyle}{
    backgroundcolor=\color{backcolour},   
    commentstyle=\color{codegreen},
    keywordstyle=\color{magenta},
    numberstyle=\tiny\color{codegray},
    stringstyle=\color{codepurple},
    basicstyle=\scriptsize\ttfamily,
    ndkeywordstyle=\color{darkgray}\bfseries,
    identifierstyle=\color{black},
    breakatwhitespace=false,         
    breaklines=true,                 
    captionpos=b,                    
    keepspaces=true,                 
    numbers=left,                    
    numbersep=5pt,                  
    showspaces=false,                
    showstringspaces=false,
    showtabs=false,                  
    tabsize=2,
}
\newcommand{\mytitle}{
Bifurcated Attention: Accelerating Massively Parallel Decoding with Shared Prefixes in LLMs
}

\icmltitlerunning{\mytitle}

\begin{document}

\twocolumn[
\icmltitle{\mytitle}



\icmlsetsymbol{equal}{*}

\begin{icmlauthorlist}
\icmlauthor{Ben Athiwaratkun}{equal,together}
\icmlauthor{Sujan Kumar Gonugondla}{equal,awsai}
\icmlauthor{Sanjay Krishna Gouda}{awsai}
\icmlauthor{Haifeng Qian}{awsai}
\icmlauthor{Hantian Ding}{awsai}
\icmlauthor{Qing Sun}{awsai}
\icmlauthor{Jun Wang}{awsai}
\icmlauthor{Jiacheng Guo}{awsai}
\icmlauthor{Liangfu Chen}{awsai}
\icmlauthor{Parminder Bhatia}{ge}
\icmlauthor{Ramesh Nallapati}{amazonagi}
\icmlauthor{Sudipta Sengupta}{awsai}
\icmlauthor{Bing Xiang}{gs}
\end{icmlauthorlist}

\icmlaffiliation{together}{Together.ai (work conducted at AWS)}
\icmlaffiliation{awsai}{AWS NGDE Science}
\icmlaffiliation{gs}{Goldman Sachs (work conducted at AWS)}
\icmlaffiliation{amazonagi}{Amazon AGI (work conducted at AWS)}
\icmlaffiliation{ge}{GE HealthCare (work conducted at AWS)}
\icmlcorrespondingauthor{Ben Athiwaratkun}{ben.athiwaratkun@gmail.com}
\icmlcorrespondingauthor{Sujan Kumar Gonugondla}{gsujan@amazon.com}

\icmlkeywords{Machine Learning, ICML}

\vskip 0.3in
]



\printAffiliationsAndNotice{\icmlEqualContribution}  


\begin{abstract}

This study introduces bifurcated attention, a method designed to enhance language model inference in shared-context batch decoding scenarios. Our approach addresses the challenge of redundant memory IO costs, a critical factor contributing to latency in high batch sizes and extended context lengths. Bifurcated attention achieves this by strategically dividing the attention mechanism during incremental decoding into two separate GEMM operations: one focusing on the KV cache from prefill, and another on the decoding process itself.
While maintaining the computational load (FLOPs) of standard attention mechanisms, bifurcated attention ensures precise computation with significantly reduced memory IO. 
Our empirical results show over 2.1$\times$ speedup when sampling 16 output sequences and more than 6.2$\times$ speedup when sampling 32 sequences at context lengths exceeding 8k tokens on a 7B model that uses multi-head attention.
The efficiency gains from bifurcated attention translate into lower latency, making it particularly suitable for real-time applications. For instance, it enables massively parallel answer generation without substantially increasing latency, thus enhancing performance when integrated with post-processing techniques such as re-ranking.

\end{abstract}

\section{Introduction} \label{sec:intro}

The advent of large language models (LLMs) has ushered in a new era of machine learning, exhibiting remarkable performance on a wide array of tasks \citep{gpt3, gpt4, palm, llama, codex, chinchilla, alphacode, copilot, cw, codegen2}. Despite their impressive capabilities, the deployment of these large-scale models in practical applications poses significant challenges, particularly in terms of inference latency and efficiency. Enhancing these aspects is critical, as they directly influence the computational resources required to generate predictions and enable the practical implementation of these advanced models across various industries. 


A particularly demanding inference scenario is single-context batch sampling, where the goal is to generate multiple completions from a single context. This task is commonly encountered in numerous applications such as code-editing IDE tools that provide multiple recommendations, or in cases where ranking among many generations is needed for optimal performance (via ranking metrics like mean log probability, majority voting, etc). The incremental decoding of such sampling scenario is memory IO intensive, which becomes a latency bottleneck for high batches and context lengths.

In this study, we investigate two compatible strategies to address the memory IO challenges in tranformers inference: (1) an investigation of multi-query and its trade-offs, and (2) a novel technique called context-aware bifurcated attention. 

Our investigation begins with an analysis of the generalized multi-query attention \citep{generalized_mq}, which includes multi-query \citep{multi_query}, as well as the established multi-head attention mechanism \citep{transformers} for performance and latency trade-off.
Our findings show smooth performance scaling with increasing model size for a fixed value of the number of groups $g$ for generalized multi-query\footnote{Lower values of attention groups $g$ lead to higher compression of the key-value tensors, as in the multi-query case where $g=1$, hence improving inference efficiency and latency due to reduced KV cache compared to the multi-head case where $g=h$, the number of query attention heads.}. Lowering $g$ results in an upward shift of the validation loss vs model size scaling curves. The consistent relationship between the cache compression, model size and validation loss allows us to trade-off inference efficiency with model size, i.e., enables us to select higher compression for use cases requiring high efficiency, while still matching the performance of multi-head attention by compensating with a larger model size. 


Secondly, we introduce context-aware bifurcated attention, a technique that bifurcates any attention in the generalized multi-query family into context and decoding components during incremental decoding.
Such bifurcation involves the same number of FLOPs and yields identical results compared to the original attention, but can significantly reduces memory IO cost and thus latency in high batch and context length scenarios.  
This approach allows the generation of multiple real-time completions without incurring much additional latency costs, or enables much higher batch sizes leading to improved ranking performance. For instance, for CodeGen 16B multi-head model \citep{codegen} with 2k context length, we are able to increase the batch size to $128$ with bifurcated attention, compared to batch size of only $5$ without, resulting in the pass@k \citep{codex} increasing from $59.0\%$ to $84.6\%$, or pass@top3 via mean log-p increasing from $55.2\%$ to $58.1\%$. 


\section{Related Work} \label{sec:related_work}

In the literature, there are multiple avenues to improve inference latency and/or latency. 
Quantization reduces memory usage by using low-bitwidth representations such as \texttt{int8}, \texttt{int4}, and \texttt{fp8} \citep{greener_yet_powerful,zero_quant,llm_int8,gptq, fp8, smoothquant}. 
Quantization when applied only to model parameters offer diminishing results as with longer sequence lengths and large batch sizes where memory access and compute associated with dot-product attention dominates the overall inference latency.

Sparse attention \citep{longformer,sparsetransformer,bigbird} has been extensively studied as a way to reduce the complexity of attention for longer contexts and faster inference.
\citet{palm_inference} investigates generative inference efficiency of large language models by using multi-dimensional partitioning techniques optimized for TPUs (collective einsum) to achieve a Pareto frontier on latency and model FLOPs utilization. The paper also shows that multi-query attention allows scaling up to 32x larger context length with an emphasis on the efficiency under high batch size. Paged attention \cite{kwon2023efficient} enhances memory management of the KV cache by dividing it into blocks and employing a block table for mapping purposes. This approach effectively accommodates dynamic workload shifts and reduces memory storage requirements through the sharing of the prompt's KV cache across multiple output sequences. However, this does not reduce the memory reads of KV cache.
%

Speculative decoding, and its variants uses a smaller draft model to propose multiple sequential tokens, which are processed in parallel by the main model to accept or reject such tokens \citep{speculative_decoding_deepmind, speculative_decoding_google, eagle, medusa, lookahead}. The key idea is to enable decoding of multiple tokens at every step, thereby amortizing the memory IO usages of the main model. However, the latency of decoding will be still dominated by KV cache I/O bandwidth at large context sizes, where bifurcated attention can enhance the decoding speed further. In short, incremental decoding focuses on lowering the amortized memory IO of model loading while multi-query and bifurcated attention lowers the memory IO of KV cache.


Additionally, we acknowledge concurrent work by \citet{juravsky2024hydragen} which presents methods to improve inference efficiency with shared-prefixes, that coincides with bifurcated attention.

\section{Background}


\subsection{Notation}
We use the following notation throughout the paper.
\begin{itemize}
\item $K$: key tensor, $V$: value tensor, $q$: query tensor, $P_x$: projection tensor associated with key, value or query tensor. 
\item We denote $\langle A,B \rangle$ as a tensor operation between $A$ and $B$. The actual operation can be specified in Einstein sum notation. We use $\oplus$ to denote concatenation.
\item $N$ the number of model parameters, $d$: hidden dimension, $h$: number of attention heads, $k$: $\frac{d}{h}$, or head dimension, $\ell$: number of layers, $m$: context length (or key/value tensor length), $n$: query tensor length where $n=m$ during context encoding and $n=1$ for incremental decoding, $g$: number of attention groups (to be explained). We also use $v$ to represent the head dimension for the value tensor where practically $k=v$. 
\end{itemize}

\subsection{Language Model Inference} \label{sec:background}

\begin{figure*}[t]
\centering
\includegraphics[trim=120 210 120 185, clip, width=0.75\textwidth]{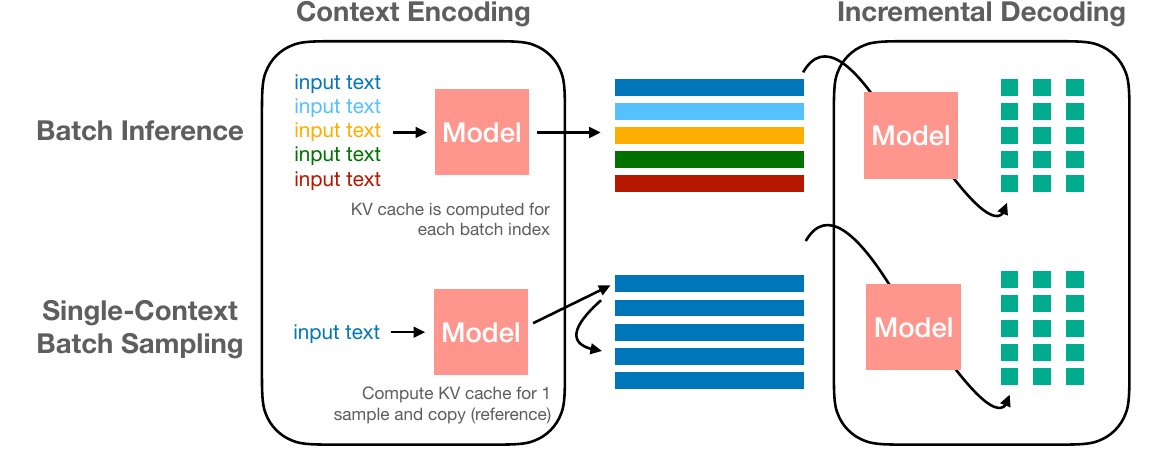} 

\caption{Illustration of the two phases of language model inference: context encoding and incremental decoding, as well as different inference scenarios.
In batch inference scenario, we process multiple inputs at once and perform incremental decoding steps. 
In batch inference, we group multiple inputs in batch to perform both context encoding and the subsequent incremental decoding. 
In the single-context batch sampling scenario, we perform context encoding on a single input to obtain the context KV cache, then perform incremental decoding (with temperature sampling) to obtain potentially different generations.
}
\label{fig:model_inference}
\label{fig:inference_scenarios}
\end{figure*}

There are many inference scenarios for language model, including batch inference and single-context batch sampling (Figure \ref{fig:model_inference}). Batch inference refers to the case where we process multiple inputs together in a batch, and generate subsequent tokens for each batch index independently. In the case where the batch size is 1, this reduces to the single-context inference. Another scenario is the single-context batch sampling where we generates multiple sequences based on a single context, where difference between the batch inference case is that the prefill only needs to be done for a single context to obtain the KV cache, then broadcasted to other batch indices.

Figure \ref{fig:model_inference} also illustrates the two phases of language model inference: (a) the context encoding or prefilling and (b) the incremental decoding. The context encoding refers to a single forward pass that computes the key and value tensors for all token positions in the context. Once the key and value tensors are computed, we cache these key and value tensors to be used for the attention mechanism during the incremental decoding phase, which sequentially generates one token at a time\footnote{Or $k$ tokens at a time, in case of speculative decoding \citep{speculative_decoding_deepmind, speculative_decoding_google}}.

During the context encoding phase, the number of floating point operations relative to the memory input/output (IO) operations is high, corresponding to the compute-bound regime where the latency is influenced by the FLOPs. However, during incremental decoding where we perform attention on a single query token, this falls into a memory-bound regime where the number of computation per memory access is roughly 1-to-1 (see Appendix \ref{appendix:memory_access_analysis} for details).
The memory IO refers to the read and write operations from the high bandwidth memory (HBM) \citep{hbm} to the fast on-chip SRAM where the actual computation happens. 
The memory IO of the incremental decoding itself consists of two components: (1) the model parameter loading and (2) KV cache loading. Component (1) is constant regardless of the context length $m$ or batch size $b$ where component (2) depends on both $m$ and $b$ and dominate the overall memory IO if $m$ or $b$ are high, which can become a significant bottleneck for inference. Our work primarily focuses on reducing component (2). 




\subsection{Multi-Query, Multi-Head and the Generalized Multi-Query Attention}

Multi-query attention, proposed by \citet{multi_query}, is an attention mechanism for transformers models that uses a single head for the key and value tensors, compared to $h$ heads in the traditional multi-head attention \citep{transformers}.
This technique effectively reduces the KV memory IO by $h$ times, which leads to higher inference efficiency during incremental decoding.
 In effect, the single-head key or value tensor is shared and used to attend to all the multi-head query, hence the name multi-query. This corresponds to a compression in representation power of the key and value tensor, which we will see in the scaling laws study (Section \ref{sec:scaling_laws}) that it results in a reduced expressiveness in terms of model parameter efficiency. Such reduced expressiveness can be compensated by scaling the model bigger than the multi-head counterpart to match the representation power.

We can also extrapolate these insights to a generalized multi-query attention mechanism \citep{generalized_mq}, which provides a framework to understand both multi-query and multi-head attention, and everything in between. Here, the degree of KV  compression is dictated by the number of attention groups $g$, 
where we alternatively refer to the generalized multi-query as multi-group. 
Each attention group can be interpreted as the broadcasted attention between a single head of key or value tensor, and multiple heads of query.

In this paradigm, multi-query attention is a special case where the number of groups $g = 1$; that is, there is exactly one such group. Conversely, multi-head attention is another special case where the number of attention groups matches the number of heads ($g=h$), in which case each head in the key or value tensor attends to one head in the query.
More generally, the number of groups $g$ can lie anywhere between $1$ and $h$, indicating various degrees of compression. For practical purposes, it is most convenient when $g$ divides $h$. The attention mechanism in this setting can be expressed in terms of Einstein summation as:

\vspace{-\baselineskip}
\begin{align}
\mathrm{logits} = \langle q, K \rangle &: \text{einsum}(bgpnk, bgmk) \to bgpnm \label{eq:multi_group_k} \\
o = \langle w, V \rangle &: \text{einsum}(bgpmn, bgmv) \to bgpnv \label{eq:multi_group_v} 
\end{align}

where $p = \frac{h}{g}$ represents the attention group size. Other operations in the attention mechanism are analogous, as detailed in Appendix \ref{appendix:multigroup_details}. The memory IO complexity for the multi-query attention becomes $bgmk$ compared to $bhmk$ in the multi-head setting, a reduction by a factor of $\frac{h}{g}$ times. The FLOPs, however, are $bgpnmk = bdnm$, independent of the compression $g$, implying that in the compute-bound scenario of context encoding, the latency would be quite similar among  multi-group models of different $g$'s, including between $g=1$ and $g=h$. 

This generalized multi-group attention mechanism thus provides a unified perspective on the design space of attention architectures. By adjusting the number of attention groups $g$, one can flexibly tune these trade-offs, potentially yielding new regimes of performance for transformer models. In Section \ref{sec:capabilities_comparison}, we will look into such capability vs latency trade-off.

\section{Context-Aware Bifurcated Attention} \label{sec:bifurcated_attention}

In this section, we present a novel \textit{context-aware bifurcated attention} method that aims to reduce the memory IO cost during incremental decoding by efficiently handling the computation of attention for shared context across samples, as shown in Figure \ref{fig:bifurcated_attention}.

 
\subsection{Motivation}

\begin{figure*}[ht]
\vspace{-0.0cm}
\centering
\includegraphics[trim=20 8 20 20, clip, width=1.0\textwidth]{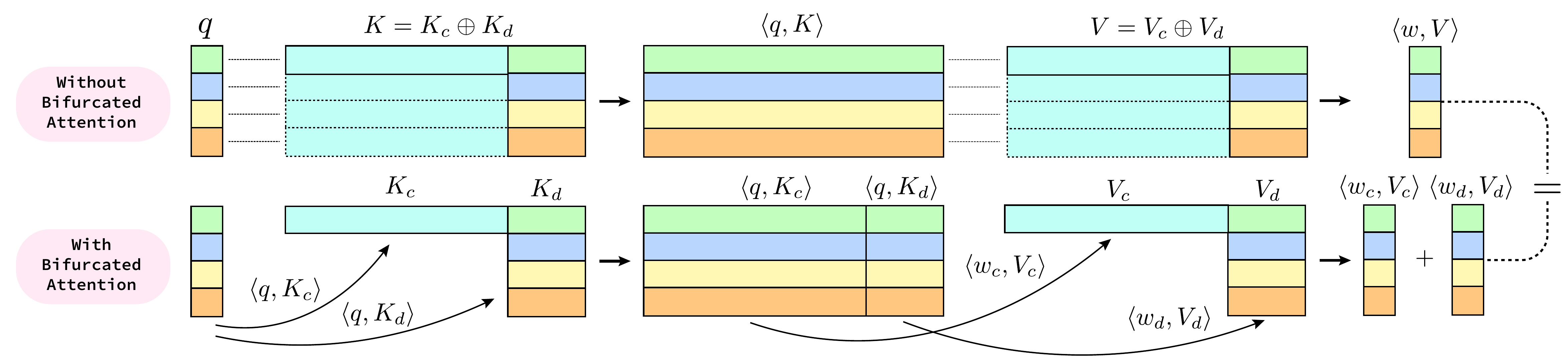}
\caption{Context-aware bifurcated attention for single-context batch sampling.
The figure depicts the incremental decoding step where the batched query $q$ attends with the cached key tensor $K$ where different colors in the $q$ tensor correspond to different batch indices. The key tensor consists of two parts: key cache corresponding to the single context $K_c$ (which was computed during context encoding, as in Figure \ref{fig:model_inference}), and the key cache corresponding to previous incremental decoding steps $K_d$. 
The query-key attention is bifurcated into two parts, $\langle q,K_{c} \rangle$ and $\langle q, K_{d} \rangle$, and joined back via concatenation, resulting in an \textbf{identical} results using the \textbf{same FLOPs} but with \textbf{lower memory IO} (Eq. \ref{eq:bifurcated_k}). The weight-value attention is bifurcated similarly, as outlined in Eq. \ref{eq:bifurcated_v}.
}

%
%

\label{fig:bifurcated_attention}
\end{figure*}

We observe that the memory IO during the incremental decoding phase can be significantly improved due to the fact that the KV corresponding to the context are shared and can be loaded only once. 
During incremental decoding, the accumulated key tensor ($K$) for a multi-head model is of size $bhmk = bh(m_c+m_d)k$. 
The two parts of $K$ correspond to $K_c$ of size $bhm_ck$ and $K_d$ of size $bhm_dk$ where $m_c$ is length of the original input and $m_d$ is the length due to previous incremental decoding steps. 
Since tensor $K_c$ is the same across all indices in the $b$ axis, we can also represent $K_c$ with a more compact shape $1h m_c k$ or simply $hm_ck$.
The query-key attention (Equation \ref{eq:multi_group_k}) is typically performed by accessing different batch indices of $K = K_c \oplus K_d$ separately, even though all batch indices in $K_c$ correspond to the same attention values. That is, if we ``naively'' pass the entire tensor to the GEMM/BLAS operators, the incurred memory I/O cost = $bhmk$, meaning that $K_c$ tensor is loaded $b$ times (Figure \ref{fig:bifurcated_attention}). Since memory loading of $KV$ is the bottleneck for incremental decoding, reducing such IO can bring significant reductions in latency saving.


\subsection{Formulation}
Below outlines the proposed context-aware bifurcated attention for single-context batch sampling. This operation splits any attention in the multi-group family during incremental decoding into two parts: (1) attention associated with KV cache from the single context $\langle q, K_c \rangle$ and (2) attention associated with KV cache from prior incremental decoding steps $\langle q, K_d \rangle $. That is, 

\vspace{-\baselineskip}
\begin{align}
\langle q, K \rangle &=  \langle q, K_c \rangle \oplus \langle q, K_d \rangle  \label{eq:bifurcated_k}\\
\langle q, K_c \rangle &: \text{einsum}(bgpnk, gm_ck) \to bgpnm_c \nonumber \\
\langle q, K_d \rangle &: \text{einsum}(bgpnk, bgm_dk) \to bgpnm_d \nonumber
\end{align}

The context part computes attention with $K_c$ that corresponds to any batch index, since they are all identical. Hence, the axis $b$ does not appears in the einsum for $\langle q, K_c \rangle$.
The result $\langle q, K_c \rangle$ and $\langle q, K_d \rangle$ are then joined together via concatenation. 
The weight-value attention $\langle w,V \rangle$ is bifurcated similarly, where the weight and value tensors are split along length $m$, and the results are joined back via summation (Eq. \ref{eq:bifurcated_v}). We also demonstrate the code for bifurcated attention in Appendix \ref{sec:bifurcated_attention_code}.

\vspace{-\baselineskip}
\begin{align}
\langle w, V \rangle &=  \langle w_c, V_c \rangle + \langle w_d, V_d \rangle    \label{eq:bifurcated_v}\\
\langle w_c, V_c \rangle &: \text{einsum}(bgpnm_c, gm_ck) \to bgpnk = bnd \nonumber \\
\langle w_d, V_d \rangle &: \text{einsum}(bgpnm_d, bgm_dk) \to bgpnk = bnd \nonumber 
\end{align}


The proposed operations yield the exact same results $\langle w, V \rangle$ as the original attention in Equation \ref{eq:multi_group_k} and \ref{eq:multi_group_v}, 
but can significantly reduce memory I/O during incremental decoding (proof in Appendix \ref{appendix:bifurcated_proof}).


\subsection{Memory IO Complexity}

The memory IO complexity corresponding to loading KV changes from

\vspace{-\baselineskip} 
\begin{align}
\text{memory IO w/o bifurcated attention} &= gk \cdot bm \\ &= gk \cdot b( m_c + m_d) \nonumber \\
\text{memory IO w. bifurcated attention} &= gk \cdot (m_c + bm_d)  
 \end{align}

The new memory IO is more efficient since
$
m_c + bm_d <  b(m_c + m_d) = bm.
$
This resulting efficiency gain  is applicable for all values of $g$ and can be as high as $b$-fold in the case where $m_c >> m_d$ (high context length compared to the number of generated tokens). 
The absolute efficiency gain, however, is more substantially for high $g$ such as in the multi-head attention case with $g=h$.
For multi-query ($g=1$), the gain can be substantial as well in the case of high $m_c$ or $b$. 



\section{Experiments}

We first conduct experiments to see how capabilities scale with respect to model size for each attention type in Section \ref{sec:scaling_laws}. We find that attention types with higher compression (lower number of attention groups $g$) require model size compensation, $\approx 10\%$ for multi-query ($g=1)$. We use such findings to compare the latency between the multi-head and the larger multi-query models of equal capabilities in Section \ref{sec:compare_two_equal}. In Section \ref{sec:bifurcated_latency}, we focus on the single-context batch sampling scenario where we demonstrate the significant latency reduction of bifurcated attention and revisit the comparison between multi-head and multi-query in light of bifurcated attention. We outline inference details in Appendix \ref{appendix:inference_hardware}. 

\subsection{Comparing Capabilities of Multi-Head, Multi-Query, and Multi-Group Attention} \label{sec:scaling_laws} \label{sec:capabilities_comparison}

For a given model configuration, a multi-group model with $g < h$ has fewer parameters in comparison to its multi-head counterpart. This reduction is a result of the decreased size of the key and value projection matrices $P_K$ and $P_V$. Specifically, each tensor in this case has a size of $P_K : d \times gk$, where $k$ is the head dimension.
For instance, a 13B multi-head model will correspond to a 11B multi-query model, with all other model configurations fixed (see Appendix \ref{appendix:memory_access_analysis} for more details).

\begin{figure}[t]
\vspace{-0.0cm}
\centering
\newcommand{\plotwidth}{0.4\textwidth}
\begin{subfigure}[t]{\plotwidth}
\includegraphics[trim=12 10 10 10, clip, width=\textwidth]{
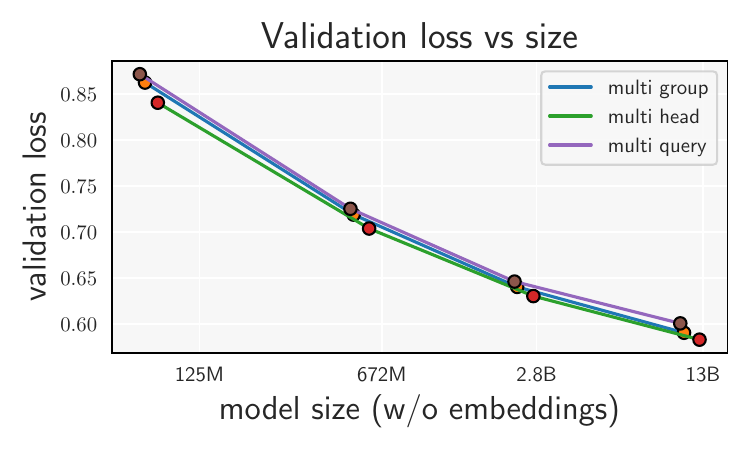
}
\end{subfigure}
\hspace{0.0cm}
\begin{subfigure}[t]{\plotwidth}
\includegraphics[trim=12 10 10 10, clip, width=\textwidth]{
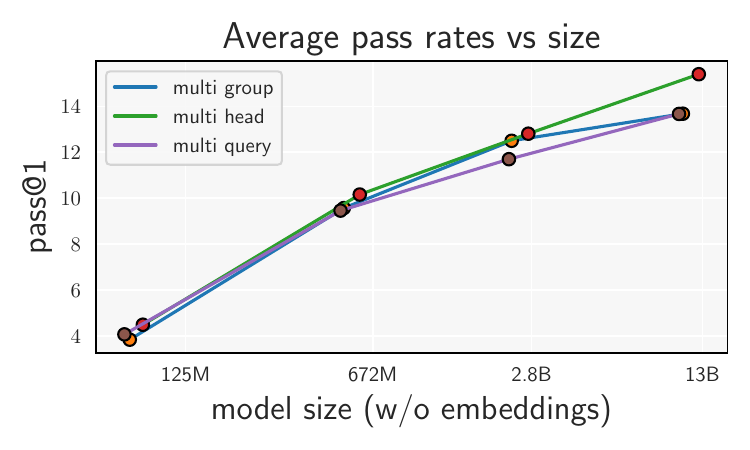
}
\end{subfigure}
\caption{(Left) The plots of validation loss versus model size  demonstrate that the scaling laws curves of different attention mechanisms have different expressiveness or performance efficiency. That is, the capabilities given the same model size depends on $g$ where higher $g$ yields the best capabilities. 
(Right) We demonstrate a similar trend where we use code generation abilities as a proxy for general capabilities. 
Here, we average the execution pass rates evaluated on Multi-lingual HumanEval and MBXP benchmarks under 13 programming languages.
%
}
\label{fig:multigroup_expressiveness}
\end{figure}

To compare the capabilities of different attention mechanisms, one can either scale other model configurations such as the number of layers $\ell$, the number of heads $h$ in order to make match the total model sizes between different attentions. However, it is often difficult to match the number of parameters exactly. In this work, we compare different attention mechanisms via the loss-vs-size scaling laws. For the setup, we use the model hyperparameters similar to that of GPT-3, where the size ranges from $125$M to $13$B, with hyperparameters such as $\ell, h, k$ increasing in tandem. Then, we consider three cases where $g = 1$ (multi-query), $g=h$ (multi-head) and $1 < g < h$ (multi-group) where Appendix \ref{appendix:training_details} and \ref{appendix:model_configurations} shows the training and model configuration details. We train all three attention models of each size and plot the validation loss versus model size, shown in Figure \ref{fig:multigroup_expressiveness}. Our findings are summarized below.

\paragraph{Higher number of attention groups $g$ leads to higher expressiveness} The results in Figure \ref{fig:multigroup_expressiveness} shows the validation loss versus model size (log scale).
The results 
 indicate that, for the same model size (vertical slice across the plot), multi-head attention $g=h$ achieves the lowest validation loss compared to $1<g<h$ (multi-group) and $g=1$ (multi-query). This trend holds consistently over three orders of magnitude of model sizes, where the curves corresponding to multi-head, multi-group and multi-query do not cross, implying that the rank of model expressiveness, or relative capabilities per number of parameters, is quite stable. An intuitive explanation is that the lower $g$ corresponds to a lower rank representation of the key and value tensors, which encodes lower representation power of the past context and therefore yields lower capabilities than higher $g$, given the same model size. 

\paragraph{Scaling laws via downstream performance} We use the average scores from two code generation benchmarks, multi-lingual HumanEval and MBXP \citep{mxeval}, as a proxy for model capabilities in addition to the validation loss.
This approach is similar to that of the GPT-4 technical report \citep{gpt4} where HumanEval (Python) \citep{codex} is used to track the performance across multiple magnitudes of compute. 
In our case, we average across all 13 evaluation languages and two benchmarks to obtain a more stable proxy for capabilities. The result in Figure \ref{fig:multigroup_expressiveness} demonstrates similar trend compared to the validation loss where the pass rate curves indicate the same relative expressiveness for multi-head, multi-group and multi-query attention.

\paragraph{Matching capabilities by model size compensation} 
Given the same capabilities (horizontal slice of the plot in Figure \ref{fig:multigroup_expressiveness}), the distance between two curves indicates the model size difference that the lower-rank attention needs to compensate in order to match the multi-head model performance. Empirically, we average the distance along the interpolated lines (log scale) and find this to correspond to $1.104$ times; that is, a multi-query model can have the same capabilities as the multi-head model if the size is increased by $\approx 10\%$ of the multi-head model size. Similarly, the gap is $ < 10\%$ for multi-group attention. Alternatively, one can argue that a multi-query model of the same size could match a multi-head if the multi-query model is given more compute. However, in the regime where we train language models until or close to convergence and the performance saturates with respect to compute, the difference in capabilities will likely remain. Therefore, the size compensation is likely the most fair approach for comparison.
%



\subsection{Latencies of Capabilities-Equivalent Models} \label{sec:compare_two_equal}

As detailed in Section \ref{sec:scaling_laws}, we've observed that an increase in the multi-query model's size is required for it to match the performance of a multi-head model. In this section, we focus on examining the latency trade-offs across diverse scenarios with both multi-query and multi-head models of similar performance capabilities. For these latency experiments, we utilize two models, each with an approximate size of 1 billion: a multi-head model and a multi-query model (detailed information can be found in \ref{appendix:details_model_latency_study}). The multi-query model chosen for these studies is larger by a multiplicative factor $F$, where $F = 1.1$.

\begin{figure}[ht]
\vspace{-0.0cm}
\centering
\includegraphics[trim=170 320 220 110, clip, width=0.5\textwidth]{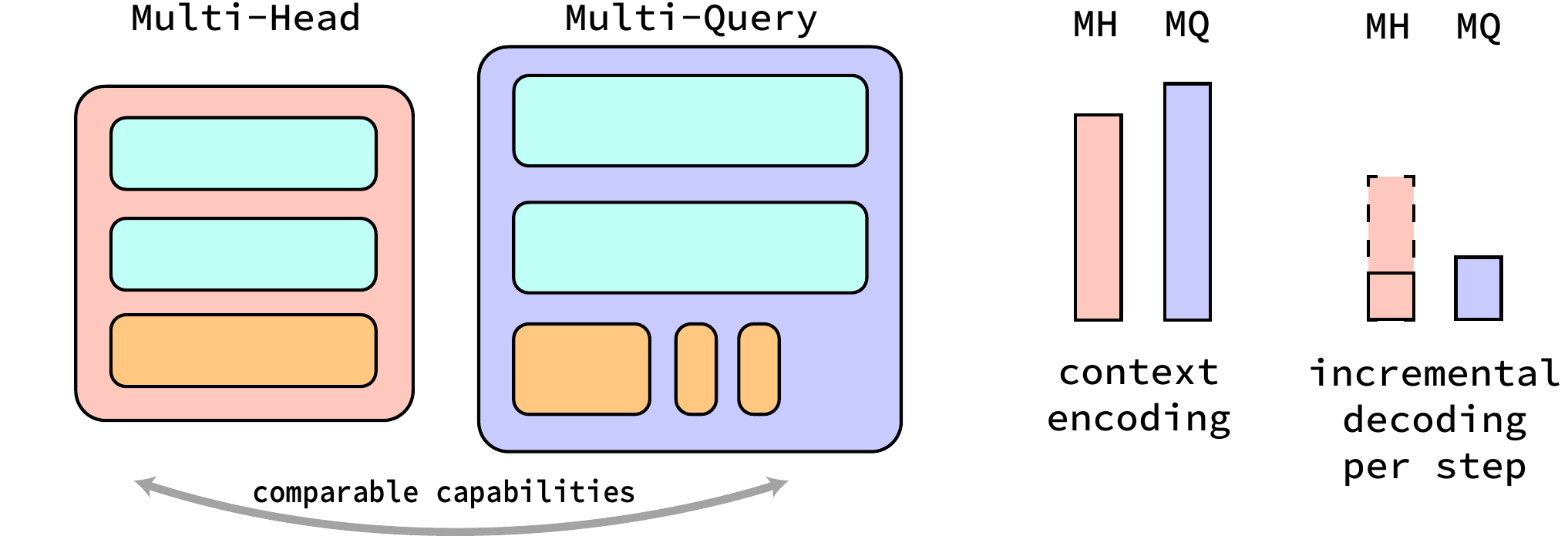}
\caption{High-level latency comparison between an MH model and a larger MQ model with comparable capabilities. 
Overall, there's an overhead cost for the initial context encoding latency due the additional compute with the larger MQ model size. For low context and batch size, the per step latency of MQ is also slightly higher to start due to the memory IO required for larger model size, but does not change much as context length $m$ or batch size $b$ grow, as supposed to the multi-head case where the per step latency can grow more rapidly with respect to $m$ and $b$. 
}
\label{fig:compare_latency_mh_mq_same_capabilities}
\end{figure}

Overall, there is some overhead cost of using multi-query attention due to the larger size (see Figure \ref{fig:compare_latency_mh_mq_same_capabilities} and Appendix \ref{appendix:compare_context_encoding_growth} and \ref{appendix:compare_incremental_decoding} for analysis). 
That is, context encoding latency of the multi-query model will be slightly larger, as well as the low-context and low-batch incremental decoding scenario. 
However, multi-query can have significantly lower latency compared to multi-head in the scenario with high number of decoding steps which makes the incremental decoding phase being latency-dominating, and high context or batch size which heavily impacts the memory  IO of incremental decoding. We outline three different inference scenarios below.

\subsubsection{Single Context Scenario} \label{sec:single_batch_scenarios}

\begin{figure*}[h]
\centering
\newcommand{\plotwidth}{0.99\textwidth}

\begin{subfigure}[t]{\plotwidth}
\includegraphics[trim=10 10 10 35, clip, width=\textwidth]{
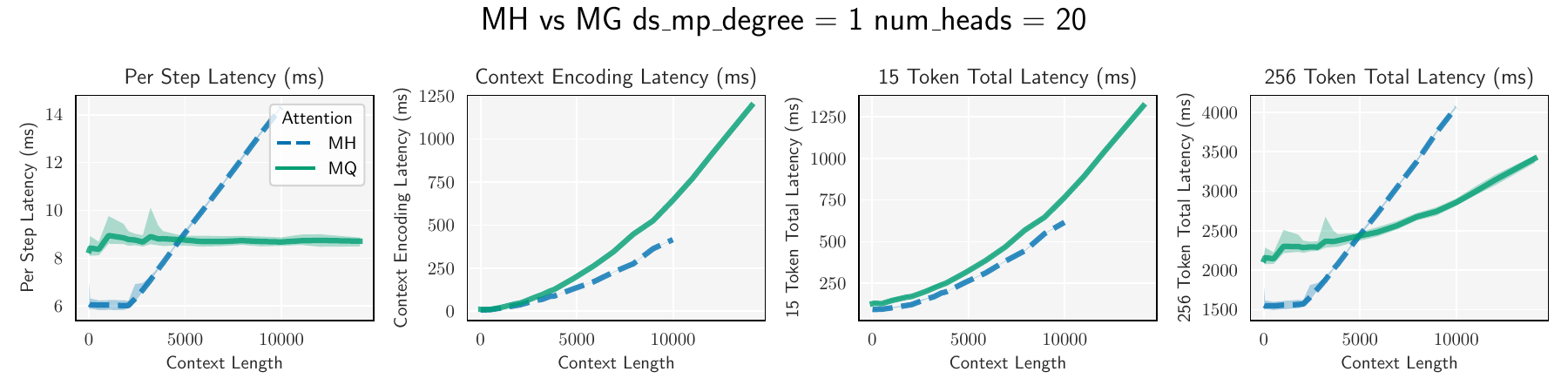
}
\end{subfigure}
\caption{
Incremental decoding (per step) latency and the context encoding latency, as a function of input context length. In this plot, we compare an multi-head model and an multi-query model of comparable capabilities, whose size is slightly larger. 
 \textbf{(Leftmost: Per-step incremental decoding latency)} 
 For low context length such as $m < 2500$, due to the larger size of the MQ model, the inference latency is higher. However, the growth with respect to context length of the MQ model is much lower (almost flat), resulting in lower per step latency when the context length is high. 
 \textbf{(Second: Context encoding latency)} The context encoding latency depends on the FLOPs where the MH and MQ are quite similar. Note that the MQ model is slightly larger, and therefore corresponds to a steeper curve.
 \textbf{(Third, Fourth): Total latency for 15 or 256 generated steps} The two plots illustrates the \emph{total} latency, which is the sum of context encoding and the the number of steps times incremental decoding latency. The benefits of MQ model becomes clear in the case of high decoding steps $(256)$ whereas in the case of $15$ generated tokens, the total latency of MQ can still be slightly higher than MH. 
}
\label{fig:single_context}
\end{figure*}

In the single batch inference scenario, the multi-query/-group attention can achieve lower latency when the context length and the number of generated tokens are high, as demonstrated in Figure \ref{fig:single_context}. 
Different implementations that are more efficient in loading KV cache (such as lower-level kernel that can avoid duplicated IO) can cause the overall curves of MH to be flatter. However, the overall trend still remains where given sufficiently high context $m$, MQ will begin to be faster than MH.

\begin{figure*}[ht]
\centering
\newcommand{\plotwidth}{0.22\textwidth}

\begin{subfigure}[t]{\plotwidth}
\includegraphics[trim=10 10 765 25, clip, width=\textwidth]{
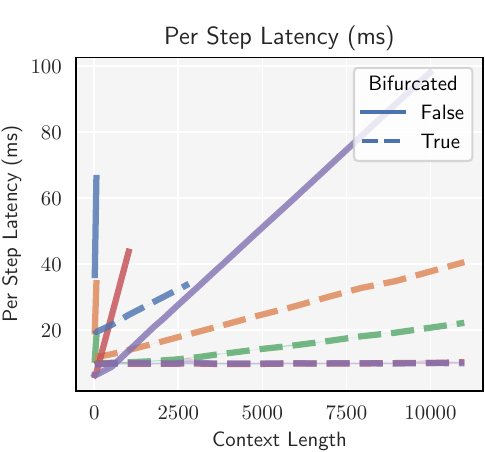
}
\caption{Multi-Head} \label{fig:bifurcated_mh}
\end{subfigure}
\hspace{0.1cm}
%
\begin{subfigure}[t]{\plotwidth}
\includegraphics[trim=10 10 765 25, clip, width=\textwidth]{
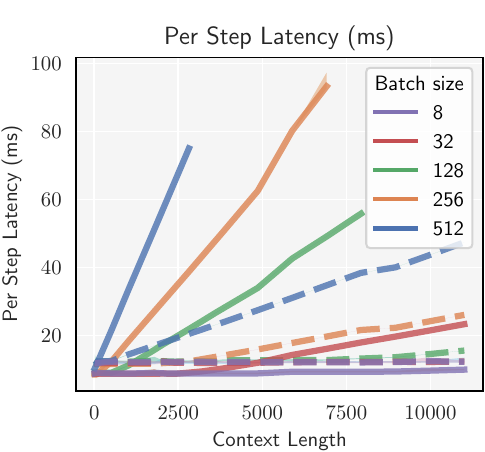
}
\caption{Multi-Query} \label{fig:bifurcated_mg}
\end{subfigure}

\caption{Context-aware bifurcated attention with multi-head attention (a) and multi-query attention (b). 
The bifurcated attention loads the KV cache in a context-aware manner, resulting in significantly lower latency for sampling under high batch sizes. For instance, in the case of multi-head attention with batch size $128$ and context length $10,000$,  bifurcated attention results in $\approx 4 \times $ lower the incremental decoding latency. Additionally, growth with respect to context length is relatively flat with bifurcated attention.
With multi-query attention, bifurcated attention permits us to use batch sizes as high as $256$ or $512$ with lower latency than in the multi-head scenario. }


\label{fig:bifurcated_mh_mg}
\end{figure*}

\subsubsection{Single-Context Batch Sampling} \label{sec:single_context_batch_sampling} \label{sec:bifurcated_latency}

In this scenario, we are given a single context and generates multiple completions based on temperature sampling. In this case, the context encoding is independent of the batch size $b$ since it is performed on the single context and broadcasted for other batch indices (Figure \ref{fig:inference_scenarios}). In contrast to the batch inference scenario, this is a more practical online inference scenario since we are not bottlenecked by the context encoding step. Our proposed context-aware bifurcated attention is exactly applicable for such scenario where in this section we demonstrate the results in conjunction with both multi-head and multi-query. 




\paragraph{Multi-head benefits significantly from bifurcated attention}

Figure \ref{fig:bifurcated_mh} demonstrates the per step latency results for a multi-head model. 
For instance, with batch size $8$, the per step latency without bifurcated attention grows rapidly with context length, from $\approx 10$ ms to $\approx 100$ ms at context length $10000$. However, with bifurcated attention, the latency remains relatively flat with respect to context length.
In practice, bifurcated attention also reduces memory consumption at high batch size and context lengths without encountering out-of-memory error as early as without bifurcated attention.

\begin{figure*}[ht]
\vspace{-0.15cm}
\centering
\newcommand{\plotwidth}{0.95\textwidth}
\begin{subfigure}[t]{\plotwidth}
\includegraphics[trim=10 10 10 25, clip, width=\textwidth]{
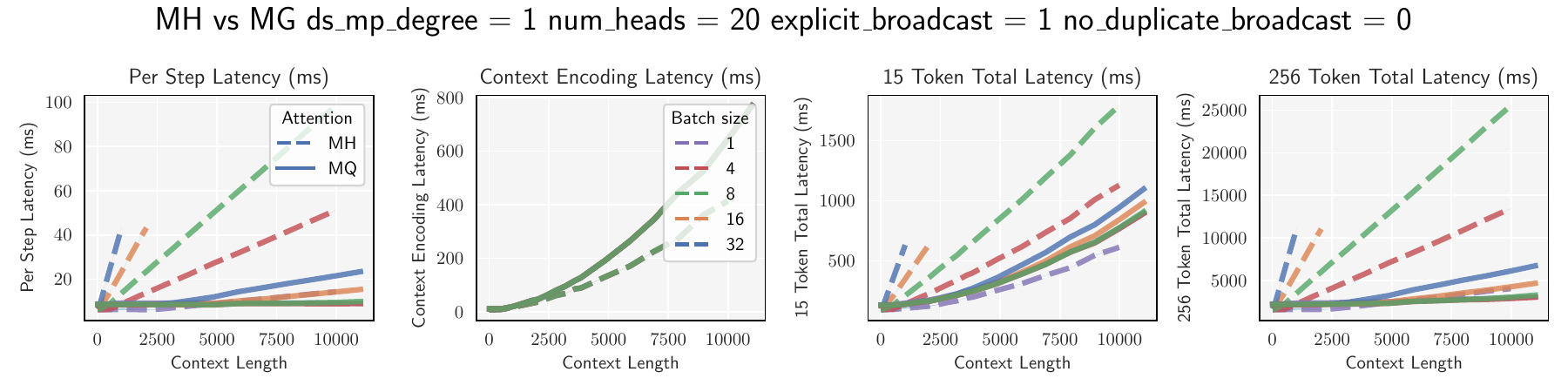
}
\caption{Without bifurcated attention}
\end{subfigure}

\begin{subfigure}[t]{\plotwidth}
\includegraphics[trim=0 10 10 25, clip, width=\textwidth]{
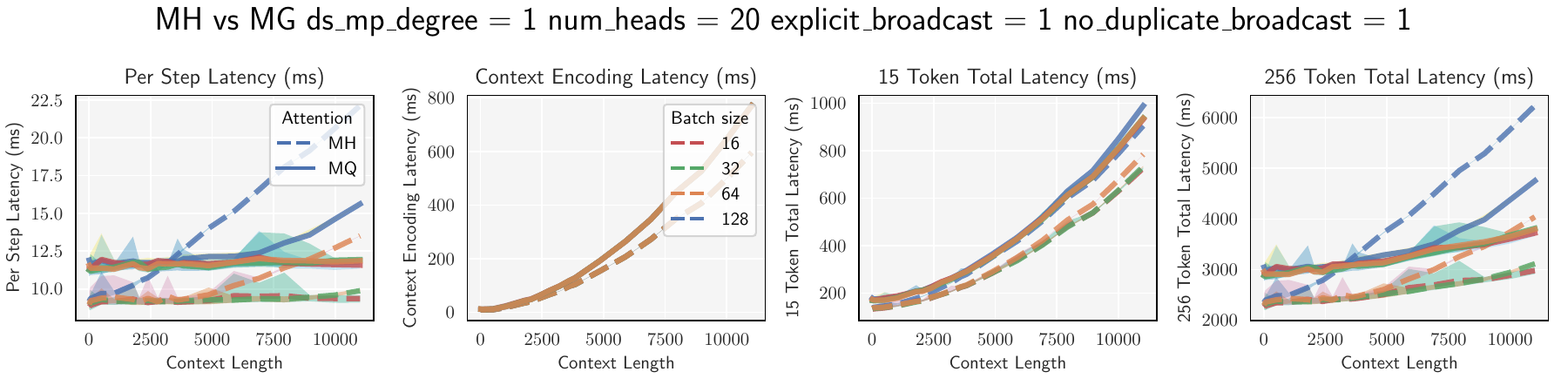
}
\caption{With bifurcated attention}
\end{subfigure}
\caption{Latency comparison between multi-head and a larger multi-query model of equal capabilities. Without bifurcated attention, MQ is clearly much more inference efficient. However, with bifurcated attention, MH can have better latency than MQ in moderate scenario (up to batch size 64 in this case) where MQ can handle more extreme scenarios better than MH. 
}
\label{fig:single_context_batch_sampling}
\end{figure*}
\begin{figure*}[ht!]
    \centering
    \includegraphics[width=0.95\textwidth]{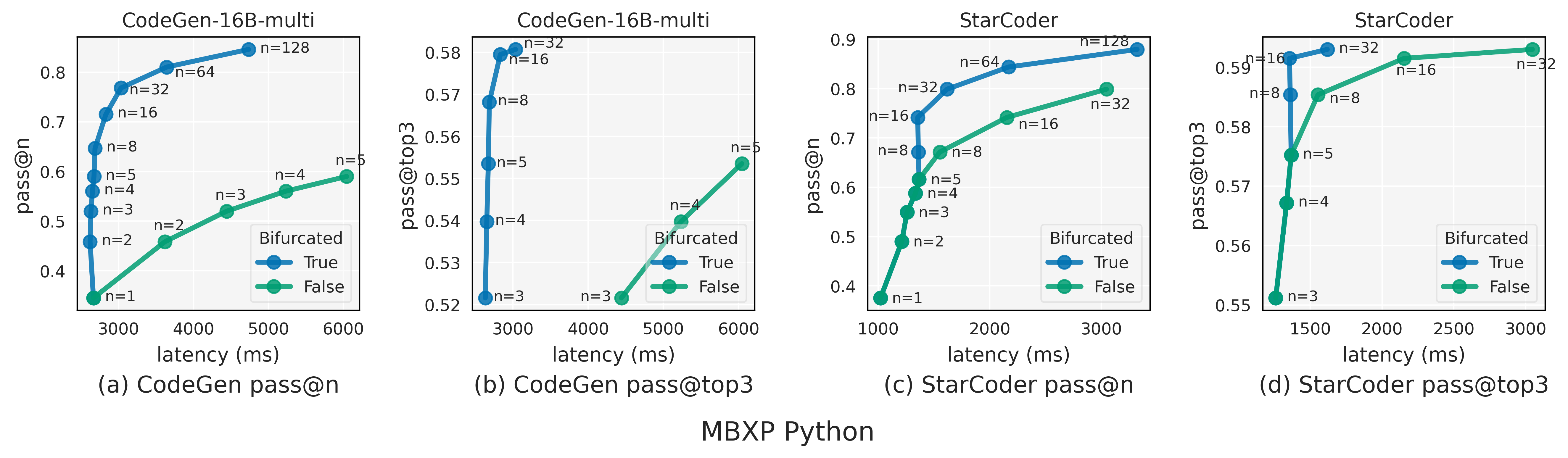}
    \caption{Bifurcated attention improves accuracy by enabling more generated samples over a fixed latency budget, applicable for both multi-head attention (CodeGen) and multi-query attention (StarCoder). 
    Given the $n$ samples, pass@n reflects the execution pass rate of the best sample among $n$, shown in (a) and (c). Filtering $n$ samples with mean log probability ranking yields a subset of best three samples, reflected by pass@top3 in (b) and (d).
    The increased number of samples within the same latency budget 
    results in increased performance via either pass@n or pass@top-k.
    }
    \label{fig:bifurcated_applications} \label{fig:application_python}
\end{figure*}

\paragraph{Bifurcated attention + multi-head rivals multi-query} 

Figure \ref{fig:single_context_batch_sampling} shows the comparison between MH and MQ with and without bifurcated attention.
Without bifurcated attention, MQ is clearly much more inference efficient. However, with bifurcated attention, MQ and MH under moderate batch size scenarios (up to 64) seems comparable, where multi-head is even has lower latency. 
The results indicate that, given an existing MH model, we can support batch sampling scenarios using bifurcated attention without the need  of a multi-query model (which requires training a new model, or at least continuous training) \citep{generalized_mq}. 
With a more inference-intensive scenarios, including batch inference scenario where the bifurcated attention is not applicable, switching to multi-query can be worth the effort.

\paragraph{Bifurcated attention with multi-query enables more extreme batch size and context lengths}

Multi-query has overall $h$ times lower memory IO and can already reduce latency for some inference scenarios. With bifurcated attention, the supported context lengths and batch sizes can become much more extreme, as demonstrated in Figure \ref{fig:bifurcated_mg}.





\subsection{Compatibility with Torch-Compile}

Bifurcated attention can be implemented with 4 einsum calls in native PyTorch, making it compatible with Torch-Compile. With Torch Compile, we can take advantage of kernel-fusion and concurrency to improve the latency of the model. To demonstrate this, we implement bifurcated attention on top of GPTFast \cite{GPTFast} \footnote{Link to our code: \href{https://github.com/bifurcated-attn-icml-2024/gpt-fast-parallel-sampling}{https://github.com/bifurcated-attn-icml-2024/gpt-fast-parallel-sampling}}.

We experiment on a 7B parameter model, with a hidden dimension of 4096 that is 32 layers deep and has 32 heads, as shown in Table~\ref{tab:torch-compile}. We observe that the overall latency imporvements fwith large-batch sampling with respect to standard SDPA attention remains consistent even when we compile the model.

\begin{table}[]
\small
\caption{Per-token latency (ms) of a 7B multi-head model on GPT-Fast with and without Torch Compilation compared to a model using Torch's standard SDPA kernel.}
\begin{tabular}{crrrrr}
\toprule
\multicolumn{1}{l}{}        &    & \multicolumn{2}{c}{ without Compile} & \multicolumn{2}{c}{Compiled}    \\
\midrule
\multicolumn{1}{l}{Context} & BS & SDPA              & Bifurcated            & SDPA           & Bifurcated     \\
\midrule
\multirow{5}{*}{8k}         & 1  & 26.40             & 30.39                 & 8.78           & 8.64           \\
                            & 2  & 28.71             & 31.37                 & 10.51          & 11.77          \\
                            & 4  & 43.36             & 31.44                 & 13.23          & 12.03          \\
                            & 8  & 72.71             & 33.72                 & 17.33          & 12.36          \\
                            & 16 & 132.89            & 31.71                 & 26.19          & 12.60          \\
\midrule
\multirow{5}{*}{16k}        & 1  & 30.13             & 30.66                  & 13.06          & 12.16                   \\
                            & 2  & 44.74             & 32.62                  & 15.35       & 17.17           \\
                            & 4  & 73.62             & 33.44                 & 20.65     & 17.33          \\
                            & 8  & 132.29            & 34.67                 & 32.06  & 18.07          \\
                            & 16 & 251.47            & 36.78                & OOM    & 18.46          \\
\midrule
\multirow{5}{*}{32k}        & 1  & 44.94             & 39.97                 & 19.80 & 20.90          \\
                            & 2  & 69.22             & 48.61                 & OOM            & 29.34 \\
                            & 4  & OOM               & 49.77                 & -              & 29.73 \\
                            & 8  & -                 & 51.31                 & -              & 30.30 \\
                            & 16 & -                 & 54.92                 & -              & 30.66
\\
\bottomrule
\label{tab:torch-compile}
\end{tabular}
\end{table}

\subsection{Applications}\label{sec:application}
Efficient large-scale sampling is particularly useful for downstream applications that require multiple generations but has latency constraints, e.g., AI code assistants. In this case, bifurcated attention enables generating more candidates by using larger batch size without incurring much additional latency. 
To verify our point, we empirically evaluate CodeGen-16B-mono \citep{codegen} and StarCoder (15.5B) \citep{starcoder} on MBPP dataset \citep{mbpp}, and plot pass rates with respect to latency in Figure \ref{fig:bifurcated_applications}, where we also indicate the batch size $n$. We consider two accuracy measurements: (1) \textit{pass@$n$} corresponds to the oracle scenario, where we evaluate all the generated samples and check if any of them is correct; (2) \textit{pass@top3}, where we are only allowed to evaluate three examples no matter how many we generate. In the top-3 case, we deduplicate the $n$ samples, and rank by their mean log probability scores \citep{codex} to determine three candidates. All experiments use nucleus sampling with $p = 0.95$ \citep{top_p} and temperature $0.8$. The results show much sharper improvement in either metrics relative to additional latency. This approach opens up avenues for performance improvement given a fixed budget of latency.

Many reasoning algorithms such as self-consistency Chain-of-thought (SC-COT) \cite{wang2023selfconsistency} and Tree-of-thought (ToT) \cite{yao2023tree} depend on sampling multiple outputs with shared prefix, where bifurcated attention will enable higher accuracy under same costs.



\section{Conclusion}

Bifurcated attention provides a complementary approach to the existing inference acceleration methods, with a particular focus on minimizing the memory IO of the incremental decoding, thereby enhancing inference efficiency. Our work helps support demanding inference scenarios due to larger context during incremental decoding, which are emerging from, e.g., more complex applications that requires long context such as complex reasoning, planning, or retrieval augmented generations.

\section*{Impact Statement}

Bifurcated attention is an approach that can significantly reduce the latency and associated costs involved in deploying large language models (LLMs). A key advantage of this technique is its potential to lower the carbon emissions associated with LLM inference. While reducing deployment costs could potentially lead to broader adoption of LLMs, the societal impact of such increased usage remains difficult to predict with certainty. 
Nonetheless, bifurcated attention presents an opportunity to make LLM deployment more efficient and environmentally friendly, although the broader implications warrant careful consideration.

\small{
\small{
\bibliographystyle{abbrvnat}
\bibliography{references}
}
}
\appendix




\section{FAQs}
\begin{enumerate}
\item 
\textbf{Q}: If we already have an MQ model that seems to be quite efficient at large batch sampling, is bifurcated attention necessary?
\\ \textbf{A}: The proposed context-aware bifurcated attention is an exact computation that provides a different way to perform attention, so one can use it "for free" without a performance tradeoff. Due to the reduced memory I/O, it enables more extreme cases of batch sampling, such as a larger batch, even for long contexts.

\item 
\textbf{Q}: How applicable is multi-query for single-batch inference without high batch sampling?
\\ \textbf{A}: If the context is long and the number of generated tokens is high, then the benefits of multi-query are clear. Please see Section \ref{sec:single_batch_scenarios}.

\item
\textbf{Q}: Is bifurcated attention applicable for the case where we process different inputs in a batch?
\\ \textbf{A}: No. In that case, if we need a solution to reduce memory I/O during incremental decoding, then multi-query attention can be appealing, especially in scenarios with a high number of generated tokens where the incremental decoding phase dominates the overall latency. This is because there is an overhead to multi-query due to the context encoding phase, as outlined in the main paper. 

\item 
\textbf{Q}: Any caveats to using bifurcated attention?
\\ \textbf{A}: For small workloads (low context length and batch size), due to the fact that we split the attention into two parts, there can be less parallelization of the GEMM kernels, which could lead to higher latency, especially for MQ models. However, one can get the best of both worlds given any model by triggering bifurcated attention under high workload scenarios and using normal attention otherwise. With such a workload-based switch, bifurcated attention is guaranteed to provide better latency and efficiency.

\item 
\textbf{Q}: How does model quantization (or lower precision arithmetic) affect the findings?
\\ \textbf{A}: There are two regimes for quantization: model weight quantization and attention quantization. To date, most quantization only focuses on the weight since the attention computation is precision-sensitive and quantization has not proved to be viable. 

Model quantization can make incremental decoding faster due to lower memory I/O of the model itself, since the effective model size in memory is smaller. This shifts the latency curve downward for all context lengths or batch sizes. The overall conclusion for the bifurcated and multi-query attention remains the same, however, since the improvement proposed in the paper is on the attention component, which is orthogonal to the model weight. 

If attention quantization is viable in the future, the lower memory on the attention tensor will effectively reduce the memory I/O for KV cache by a factor of 2 in the case of \verb|int8| quantization (compared to \verb|fp16| or \verb|bf16|) or a factor of 4 in the case of \verb|int4|. Overall, this will flatten the latency growth with respect to batch size or context length. The overall comparative complexity (a) with or without bifurcated attention or (b) multi-head vs. multi-query remains the same.

\item \textbf{Q}: Does the conclusion depend on the inference implementation or different hardware?
\\ \textbf{A}: Different inference platforms, such as \texttt{FasterTransformers (GPUs)} or \texttt{PaLM inference (TPUs)}, can yield different latency numbers. However, the relative I/O complexity among different attention mechanisms does not change, resulting in similar relative trends among different attention mechanisms. That being said, it is possible that more efficient implementations or more performant chip/system configurations, including different tensor parallelism degrees, can result in different slopes for the latency growth with respect to context length and batch size. In that case, the trade-off points in terms of context length or batch size can be different. The comparative complexity remains the same based on the analysis.

\item \textbf{Q}: How does bifurcated attention differ from using attention mask for sampling as in done in SpecInfer \citep{specinfer} ?
\\ \textbf{A}: The attention mask approach can have a different FLOP usage compared to the original attention. We can consider a scenario where the attention mask corresponds to sampling with batch $b$ and incremental decoding length $\ell$, with the original context of length $m$. The attention FLOPs are $O(m b \ell + b^2\ell^2)$. In contrast, the original FLOPs is $O(mb\ell)$. If $b\ell$ is sufficiently large, then the FLOPs via attention mask can be much higher. However, for the purpose of speculative decoding where the number of draft tokens is small, this additional FLOPs can be negligible.

\end{enumerate}

\section{Related Work} \label{appendix:related_work}

\subsection{Applications of Single-Context Batch Sampling} \label{appendix:applications}

The observed latency reduction we achieve can have a profound impact on many applications. Some of these applications include:
\begin{itemize}

\item Code Generation: In software development, AI-assisted code generation can benefit greatly from reduced latency, especially when generating multiple code snippets or suggestions for a given context. This can lead to a more responsive and efficient user experience for developers using AI-powered Integrated Development Environments (IDEs) or code completion tools \citep{codegen2, codegen, codex, coderl, incoder, alphacode, santacoder, starcoder, wasiunified}.

\item Machine Translation: In situations where multiple translations are needed for a single input, such as generating translations with varying degrees of formality or generating translations for different dialects, the context-aware bifurcated attention can provide more efficient computation, resulting in faster and more scalable machine translation services \citep{humanmt,wm21findings,fbwmt21,noisychannelreranking}.

\item 
Chatbots and Conversational AI: Conversational agents often need to generate multiple responses to handle different interpretations of a user's input or to provide multiple suggestions. The reduced latency offered by the proposed method can significantly improve the responsiveness of chatbots, leading to a more natural and fluid conversation with users \citep{bard}.

\item
Creative Content Generation: In applications like poetry, story, or advertisement generation, the ability to generate multiple variations for a given prompt is crucial. The proposed method enables more efficient generation of diverse content, making it more feasible for real-time or large-scale applications \citep{lin2021plug,mirowski2023co,mosaicml64k, wordcraft}.

\item 
Reasoning: using Self-consistency Chain-of-Thought (CoT-SC) \cite{wang2023selfconsistency} and Tree-of-Thought (ToT) \cite{yao2023tree} requires the model to sample multiple outputs with a shared prefix. Bifurcated attention will enable larger number of reasoning paths in SC-COT and larger trees in ToT at the same cost of inference.

\item
Data Augmentation: In the context of data augmentation for machine learning, generating multiple alternative examples for a given input can help improve model robustness and generalization. With the reduced latency provided by context-aware bifurcated attention, the process of generating augmented data can be made faster, enabling more efficient use of computational resources during training.
\item General Large Scale Evaluation: In addition to the aforementioned use-cases there are many niche use-cases where LLM and other open-ended generation models are explored for toxicity \citep{pplm,rtp,stereoset}, detection of vulnerable code in generations \citep{asleep}, performance improving code edit generation \citep{pie}, programming language translations \citep{pltranslation} and many others. In all of these scenarios many generations per each prompt are gathered for a deeper understanding of the models, bifurcated attention can drastically speed up the generation process in such cases.
\end{itemize}

In conclusion, the proposed context-aware bifurcated attention method can significantly reduce memory I/O cost and improve latency in various applications, leading to increased efficiency and scalability. This method has the potential to enable new use cases and enhance the user experience in numerous AI-powered systems, making them more practical for real-world deployment.

\subsection{Supporting Long Context Requires IO-Efficient Attention} \label{appendix:applications_long_context}

As language models are becoming general purpose and highly capable, the demand for language models to handle longer context sequences has grown significantly. Recently, there is an ongoing focus on models that can handle even longer context sequences \citep{scalingbeyond1m, gpt4, mosaicml64k, mosaicml64k}. As of today, GPT-4 \citep{gpt4} supports context length of 32k tokens, and MPT-7B \citep{mosaicml64k} extends it to 64k while Anthropic’s Claude \footnote{https://www.anthropic.com/index/100k-context-windows} supports as long as 100k input length.  Most recently, Bulatov et al proposed 1M token input context length for transformers. These models push the boundaries of context understanding and generation capabilities, enabling more comprehensive discourse understanding and contextually informed responses. 

This trend is driven by the need for comprehensive discourse understanding in applications like Retrieval-Augmented Generation (RAG), as well as many complex prompting methods.
Applications such as RAG \citep{guu2020retrieval, izacard2022atlas, menick2022teaching, zhen2022survey} retrieve extensive passages or documents from external corpora, providing rich and grounded context for generating responses. Additionally, models like Toolformer \citep{schick2023toolformer} and WebGPT \citep{nakano2021webgpt} leverage external tools, such as APIs and search engines, to expand the context and enhance generation.

Long context is disproportionately expensive for transformer family models because for vanilla self-attention both memory and time complexity are quadratic to the sequence length. To effectively handle longer context sequences, optimizing memory I/O and reducing computational overhead are critical. Currently, the dominant approaches to addressing this challenge have been to make the attention computation less expensive. \citet{longformer} proposed to sparsify self-attention using various attention patterns. \citet{wang2020linformer} explores low-rank approximation of self-attention. In addition to the compute bound improvements, advancements in memory-efficient attention mechanisms and techniques for reducing memory I/O will continue to propel the field forward, facilitating the handling of longer context sequences in language models. FlashAttention \citep{flash_attention} is proposed to speed up self-attention and reduce the memory footprint without any approximation. It leverages fused kernel for matrix multiplication and softmax operation which greatly reduces memory IO during training.

\section{Setup}
\subsection{Model Training Details} \label{appendix:training_details}
We trained multiple models with varying sizes, ranging from 125 million parameters to 13 billion parameters, using code data with a context size of 2048 and adjusting the per-GPU batch size and total number of steps according to the model size. For model training we used multiple p4 instances each equipped with 8 40GB Nvidia A100 GPUs per instance.

For our largest model family, the 13 billion parameter model, we used a global batch size of 1024, which approximately translates to 2 million tokens per batch. The settings for each model within each model-size family were kept consistent. The remaining training hyperparameters are summarized in the following table \ref{tab:training_specs}.

We use AdamW optimizer (\citep{adamw}) with $\beta_1 = 0.9$, $\beta_2 = 0.95$, and $\epsilon = 10^{-8}$. The warm-up steps were set to 2000, and a cosine annealing learning rate schedule was employed after reaching the peak learning rate.
The minimum learning rate was set to 10\% of the corresponding peak learning rate.
A weight decay (\citep{wtdecay}) of 0.01 and gradient clipping of 1.0 were applied to enhance training stability.
Following the approach in (\citep{megatron}), the standard deviation for random weight initialization was rescaled for larger models.
Our training pipeline
is based on PyTorch Lightning and we use bfloat16 (\citep{bf16}) and DeepSpeed (\citep{deepspeed}) for training optimization.
Finally, a random split of 0.1\% of the data was reserved as a validation set.
\begin{table*}[ht]
    \centering
    \caption{Training Hyperparameters}
    \label{tab:training_specs}
    \begin{tabular}{ccccc}
      \toprule
      Model Size & Total Training Steps & Batch Size & Compute Nodes & Max Learning Rate \\
      \midrule
      125M  & 400k  & 256  &8 & $2.5 \times 10^{-4}$  \\
      \midrule
      672M  & 200k  & 256  &8 & $2.5 \times 10^{-4}$  \\
     \midrule
      2.8B  & 200k  & 512  &16 & $1.6 \times 10^{-4}$  \\
     \midrule
      13B  & 100k  & 1024  &32 & $1.0 \times 10^{-4}$  \\
      \bottomrule
    \end{tabular}
  \end{table*}

\subsection{Model Configurations} \label{appendix:model_configurations}
For each model size we train three models with attention variations; multi head where $g=h$, multi group where $1<g<h$ and multi query where $g=1$. Additionally, for 672m and 2.8b models we train a multi group model variant where the fanout in feed forward layer is decreased from $4 \times d$ to $2 \times d$. Each model variant yields different number of total parameters therefore we group these models into family of model sizes. The detailed architectural choices for each of the model family is found in the table \ref{tab:model_specs}.
\begin{table*}[ht]
    \centering
    \caption{Model Specifications table presenting architecture details for the three variants: multi head (MH), multi query (MQ), and multi group (MG) including parameter count, number of attention groups, head dimensions, and number of layers. The additional fanout-based MG variant is described here as MG + $2 \times d$
 }
    \label{tab:model_specs}
    \begin{tabular}{*{6}{p{\dimexpr(\textwidth-24\tabcolsep)*10/52}}}
      \toprule
      Model Family & Attention Type & $groups$ & $d_{head}$ & $n_{layer}$ & $N_{params}$ (billions) \\
      \midrule
      \multirow{2}{*}{125M}  & MH   & 12  & 64 & 12  & 0.125 \\
      & MG  & 4  & 64 & 12  & 0.115 \\
                             & MQ  & 1  & 64 & 12  & 0.112 \\
      \midrule
      \multirow{2}{*}{672M}  & MH   & 20  & 72 & 24  & 0.672 \\
      & MG  & 4  & 72 & 24  & 0.592 \\
        & MG + $2 \times d $ & 4  & 72 & 24  & 0.393 \\
                             & MQ  & 1  & 72 & 24  & 0.578 \\
     \midrule
      \multirow{2}{*}{2.8B}  & MH    & 24  & 128 & 24  & 2.878 \\
      & MG   & 4  & 128 & 24  & 2.501 \\
      & MG + $2 \times d$   & 4  & 128 & 24  & 1.595 \\
                             & MQ   & 1  & 128 & 24  & 2.444 \\
     \midrule
      \multirow{2}{*}{13B}   & MH   & 40  & 128 & 40  & 12.852 \\
      & MG  & 8  & 128 & 40  & 11.174 \\
                             & MQ  & 1  & 128 & 40  & 10.807 \\
      \bottomrule
    \end{tabular}
  \end{table*}

\subsection{Model Details of 1B Latency Experiment} \label{appendix:details_model_latency_study}

In Section \ref{sec:bifurcated_latency}, we use candidate models of sizes roughly 1B to study the effect of bifurcated attention. We outline the hyperparameters of such models below.


\begin{table*}[ht]
    \centering
    \caption{
    Model Specifications for Latency Experiment in Section \ref{sec:bifurcated_latency}.
 }
    \label{tab:model_spec_1b}
    \begin{tabular}{*{6}{p{\dimexpr(\textwidth-24\tabcolsep)*10/52}}}
      \toprule
      Model Family & Attention Type & $groups$ & $d_{head}$ & $n_{layer}$ & $N_{params}$ (billions) \\
       \midrule
       \multirow{2}{*}{1B}  & MH    & 20  & 128 & 12  & 1.077 \\
       				     & MG   & 4  & 128 & 15  & 1.156  \\
                                       & MQ   & 1  & 128 & 16  &   1.193 \\
        \bottomrule
    \end{tabular}
  \end{table*}

\subsection{Ablation Studies: $2d$ Intermediate Feature Dimension} \label{appendix:fanout}

\begin{figure*}[ht]
\centering
\newcommand{\plotwidth}{0.5\textwidth}
\begin{subfigure}[t]{\plotwidth}
\includegraphics[trim=0 0 0 0, clip, width=\textwidth]{
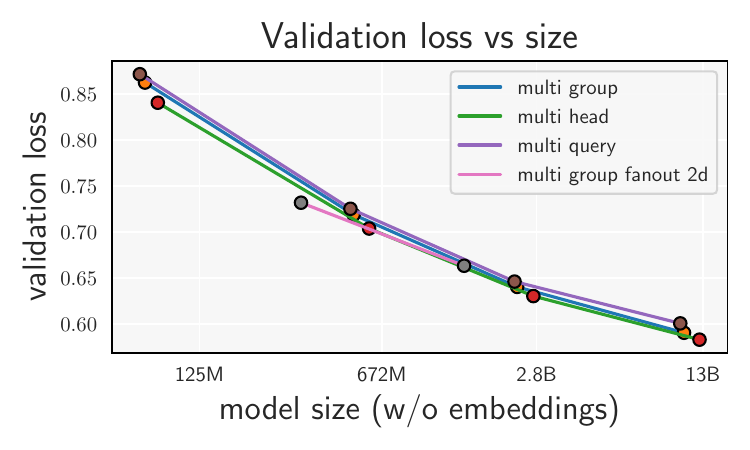
}
\end{subfigure}
\hspace{0.3cm}
\begin{subfigure}[t]{\plotwidth}
\includegraphics[trim=0 0 0 0, clip, width=\textwidth]{
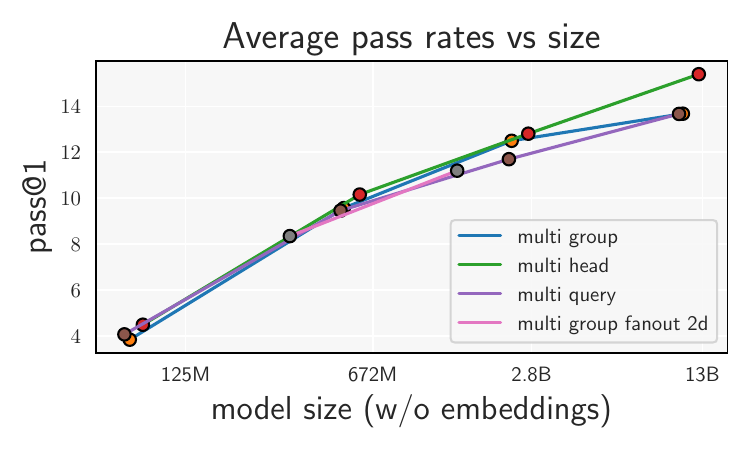
}
\end{subfigure}
\caption{
Capabilities versus size plots including the $2d$-intermediate-size feedforward model. The plot shows that the balance between the number of feedforward parameters and the attention parameters alone does not explain the relative expressiveness of multi-head, multi-group, and multi-query attentions. Rather, we argue that what explains relative expressiveness is the representation power associated with the key and value tensors (Section \ref{sec:capabilities_comparison}).
}
\label{fig:multi_group_with_fanout}
\end{figure*}

One can also argue that different $g$ results in different balance of the number of parameters in the feedforward versus the attention components. We performed an ablation study where we reduce the typical intermediate feature size of $4d$ to $2d$ and train models for three model sizes (which we will refer to as the $2d$ experiment). The ablation study reveals that the scaling laws curves for the $2d$ experiment crosses the usual $4d$ curves, which implies that the reduced size of the attention component alone compared to feedforward does not provide a consistent explanation of model capabilities. This can be seen from Figure \ref{fig:multi_group_with_fanout}.

\subsection{Inference Setup} \label{appendix:inference_hardware}
We use Nvidia A100 GPUs for inference hardware \citep{a100_gpu}. We perform latency studies using Deepspeed inference \citep{deepspeed} on top of Huggingface transformers \citep{huggingface}, where we wrote custom code to handle the generalize multi-group attention as well as bifurcated attention.  Future work includes extending the implementation to FasterTransformer \citep{fastertransformer}.

\section{Multi-Group Attention Family}
\subsection{Detailed Analysis on Memory Access} \label{appendix:memory_access_analysis} \label{appendix:multigroup_details}

We show in Table \ref{tab:comparison_attention} that the memory IO cost for $\langle q, K \rangle$ is dominated by the loading of $K$ which costs $bmhk$ in the case of multi-head where $g=h$. 
This cost is particularly high due to the coupling of batch size $b$, context length $m$, and the entire hidden dimension $d$. 
Compared to the number of computations, which has complexity $bmd$, this attention module requires one memory IO per one tensor operation (memory-io bound). In contrast, other operations such as feedforward has much lower ratio of memory IO per compute (compute bound). These attention computation can be the main bottleneck for incremental decoding and our paper aims to tackle such problems.

Concretely, we can see that the context encoding in single-batch scenario in Appendix \ref{sec:single_batch_scenarios} is $400$ ms for context length $10000$, implying that the amortized latency per token during this phase is $0.04$ ms per token. However, the per token latency during incremental decoding is in the order of $\approx 10$ ms per token, which is $\frac{10}{0.04} = 250$ times slower. This number clearly demonstrates that compute is not a dominating factor, but the memory IO required to load both model and KV cache.

\begin{table*}[ht]
\caption{Comparison of memory access and computation between Multi Head, Multi Query, and Multi Group attention mechanisms. The memory access is for incremental decoding with the query length $n=1$. 
}
\centering
\small
\scriptsize{
\begin{tabular}{l|l|c|c}
\textbf{Operation} & \textbf{Einsum} & \textbf{Memory Access} & \textbf{Computation} \\
\hline
Input ($x$): $bd$ & & & \\
$q = \langle x, P_q \rangle$ & $bd,hdk \rightarrow bhk$ & $bd + hdk = bd + d^2$ & $bdhk = bd^2$ \\
$K = \langle x, P_k \rangle \ (+ K_{prev})$ & [MH] $bd,h dk \rightarrow bhk \ (+ bmhk)$ & $bd + d^2$ & $bdhk = bd^2$ \\
 & [MQ] $bd,dk \rightarrow bk \ (+ bmk)$ & $bd + dk$ & \\
 & [MG] $bd, gdk \rightarrow bgk \ (+ bgmk)$ & $bd + gdk$ & \\
$V = \langle x, P_v \rangle \ (+ V_{prev})$ & [MH] $bd,hdv \rightarrow bhv \ (+ bmhv)$ & $bd + d^2$ & $bdhv = bd^2$ \\
 & [MQ] $bd,dv \rightarrow bv \ (+ bmv)$ & $bd + dv$ & \\
 & [MG] $bd, gdv \rightarrow bgv \ (+ bgmv)$ & $bd + gdv$ & \\
logits $= \langle q, K \rangle$ & [MH] $bhk,bhmk \rightarrow bhm$ & $bhk + bhmk = bd + bmd$ & $bhmk = bmd$ \\
 & [MQ] $bhk,bmk \rightarrow bhm$ & $bd + bmk + bhm$ & \\
 & [MG] $bhk,bgmk \rightarrow bhm$ & $bhk + bgmk + bhm$ & \\
weights: softmax & & $bhm$ & $bhm$ \\
out($O$) $= \langle$ weights, $V \rangle$ & [MH] $bhm,bhmv \rightarrow bhv$ & $bhm + bhmv = bhm + bmd$ & $bhmv = d$ \\
 & [MQ] $bhm,bmv \rightarrow bhv$ & $bhm + bmv + bhv$ & \\
 & [MG] $bhm,bgmv \rightarrow bhv$ & $bhm + bgmv + bhv$ & \\
$y = \langle O, P_O \rangle$ & $bhv,hdv \rightarrow bd$ & $bd + d^2$ & $bdhv = bd^2$ \\
\hline
Total: Multi Head & & $bd + bmd + d^2$ & $bhm + bmd + bd^2 \approx bd^2$ \\
Total: Multi Query & & $bd + bmk + d^2$ & \\
Total: Multi Group & & $bd + bgmk + d^2$ & \\
\hline
$r$: Multi Head & & $1/d + m/d + 1/b$ & \\
$r$: Multi Query & & $1/d + m/(dh) + 1/b$ & \\
$r$: Multi Group & & $1/d + g/(dh) + 1/b$ & \\
\bottomrule
\end{tabular}
}
\label{tab:comparison_attention}
\end{table*}

\subsection{Model FLOPs} \label{sec:context_encoding} \label{appendix:flops_multigroup}

The scaling laws by \citet{scaling_laws} shows that the model-related FLOPs during the forward pass is $2N$ where $N$ is the number of parameters (without the embeddings). We show that it holds for a general multi-group model as well. The only difference between the multi-group and the multi-head case is the projection $P_K$ and $P_V$ where they are of size $dgk$ instead of $dhk$. Since this is a linear layer, the forward pass FLOPs for any input is still proportional such projection size. Therefore, it follows that for any multi-group attention, including multi-head, the forward FLOPs is $2N$ where $N$ is the respective number of parameters.

\subsection{Comparing Capabilities-Equivalent Models}
This section outlines the analysis of latency change when we switch from an MH model to an MG model with $F$ times the size. 
\subsubsection{Context encoding} \label{appendix:compare_context_encoding_growth}
The dominating factor for latency in context encoding is the compute rather than the memory IO. The compute can be broken down into two parts (a) tensor projections related to model parameters and (b) KV attention involving no model parameters. For both parts, the large multi-group model will involve higher latency proportional to the size factor $F$. The context encoding time is $\propto N \times bm$ where $N$ is the model size except embeddings for (a) since the FLOPs per token is $2N$ \citep{scaling_laws}, which holds for all multi-group attention (Appendix \ref{appendix:flops_multigroup}).  For (b), the encoding time is $\propto \ell \cdot bhm^2 \propto N bm^2$ for (b). Overall, the multi-group model with similar capabilities as the multi-head model will incur slightly higher context encoding time due to the larger size since $N$ to increase to $F N$.

\subsubsection{Incremental Decoding} \label{appendix:compare_incremental_decoding}

The incremental decoding component can dominate the overall inference latency compared to the context encoding, especially in the scenario where we decode in many steps. 
Incremental decoding is memory bound, meaning that the latency of this step is limited by memory I/O throughput. We can divide the memory I/O usage into two parts: reading (a) model parameters and (b) cached key-value pairs.
With multi-group, we expect the model parameters to increase by a factor of $F(g)$, leading to an increase in I/O usage in (a) by the same factor. 
The memory IO in (b) changes by a factor of $\frac{g}{h}$ when moving from multi-head with KV cache size $2bhmk$ to multi-group with cache size $2bgmk$ (more precisely $\frac{g}{h} \cdot F(g)$ but $\frac{g}{h}$ is a dominating term since $F(g)$ is close to $1$). 

\section{Context-Aware Bifurcated Attention} \label{appendix:bifurcated_attention}

\subsection{Proof} \label{appendix:bifurcated_proof}
Here, we outline the proof that the proposed bifurcated attention in Equation \ref{eq:bifurcated_k} and \ref{eq:bifurcated_v} recovers the same attention as the operations in \ref{eq:multi_group_k} and \ref{eq:multi_group_v} for the case of single-context batch sampling. We use the fact that the KV part corresponding to context length, all the batch indices correspond to the tensors.

\begin{align*}
 \langle q, K \rangle &: \text{einsum}(bgpnk, bgmk) \to bgpnm   \\
&= \text{einsum}(bgpnk, bg(m_c + m_d)k) \to bgpnm  \\
&= \text{einsum}(bgpnk, bgm_ck) \to bgpnm \oplus  \\
 & \ \ \ \ \ \text{einsum}(bgpnk, bgm_dk) \to bgpnm \\
&= \text{einsum}(bgpnk, bgm_ck) \to bgpnm \oplus \\
& \ \ \ \ \ \text{einsum}(bgpnk, gm_dk) \to bgpnm \\
&= \langle q, K_c \rangle \oplus \langle q, K_d \rangle\\ 
\\
\langle w, V \rangle &: \text{einsum}(bgpnm, bgmk) \to bgpnk = bnd  \\
&=  \text{einsum}(bgpnm_c, bgm_ck) \to bgpnk + \\ &  \ \ \ \ \ \text{einsum}(bgpnm_d, bgm_dk) \to bgpnk   \\
&=  \text{einsum}(bgpnm_c, gm_ck) \to bgpnk + \\ & \ \ \ \ \ \text{einsum}(bgpnm_d, bgm_dk) \to bgpnk  \\
&= \langle w_c, V_c \rangle + \langle w_d, V_d \rangle
\end{align*}

\subsection{Detailed Memory I/O Analysis}

\label{sec:memory_analysis}

Overall, the memory I/O complexity changes from 
\begin{itemize}
\item Original memory I/O cost: $bhnk + bgmk + bhnm$ (for $\langle q,K \rangle$) + $bhnm + bgmk + bnd $ (for $\langle w,V \rangle$)
\item Bifurcated attention memory I/O cost: $bhnk + gm_ck + bgm_dk + bhnm$ (for $\langle q,K \rangle$) + $bhnm + gm_ck + bgm_dk + bnd$ (for $\langle w,V \rangle$)
\end{itemize}

There is an associated memory IO to write the $\langle w,V_c \rangle$ and $\langle w,V_d \rangle$ output twice. However, it is typically very small ($bnd$) compared to the IO of KV cache component $bgmk$ since $m >> n = 1$.

\subsection{Implementation of Bifurcated Attention} \label{sec:bifurcated_attention_code}
Despite the dramatic gain in inference efficiency of the bifurcated attention, we demonstrate the simplicity of our implementation involving ~20 lines of code using Pytorch \citep{pytorch}. 

\begin{lstlisting}[language=Python, xleftmargin=0.4cm]
def attn(query, key, value, bifurcated_attention):
    # <q,K>
    if bifurcated_attention and type(key) == dict:
        # g = number of groups
        # h = gp where p = num heads per group
        # n = 1 for incremental decoding
        attn_weights_context = torch.einsum(
            "bgpnk,gmk->bgpnm", query, key["context_past_key"][0]
        )
        attn_weights_incremental = torch.einsum(
            "bgpnk,bgmk->bgpnm", query, key["incremental_past_key"]
        )
        attn_weights = torch.cat(
            [attn_weights_context, attn_weights_incremental], dim=-1
        )
    else:
        attn_weights = torch.einsum(
            "bgpnk,bgmk->bgpnm", query, key
        )
    # softmax and causal mask (omitted)
    # <w,V>
     if bifurcated_attention and type(value) == dict:
        # n = 1 for incremental decoding
        context_past_value_length = value["context_past_value"].size(-2)
        attn_output_context = torch.einsum(
            "bgpnm,gmv->bgpnv",
            attn_weights[:, :, :, :, :context_past_value_length],
            value["context_past_value"][0],
        )
        attn_output_incremental = torch.einsum(
            "bgpnm,bgmv->bgpnv",
            attn_weights[:, :, :, :, context_past_value_length:],
            value["incremental_past_value"],
        )
        attn_output = attn_output_context + attn_output_incremental
    else:
        attn_output = torch.einsum(
            "bgpnm,bgmv->bgpnv", attn_weights, value
        )
    return attn_output
\end{lstlisting}

\begin{figure*}[t]
    \centering
    \newcommand{\plotwidth}{0.99\textwidth}

    \begin{subfigure}[t]{\plotwidth}
    \includegraphics[width=\textwidth]{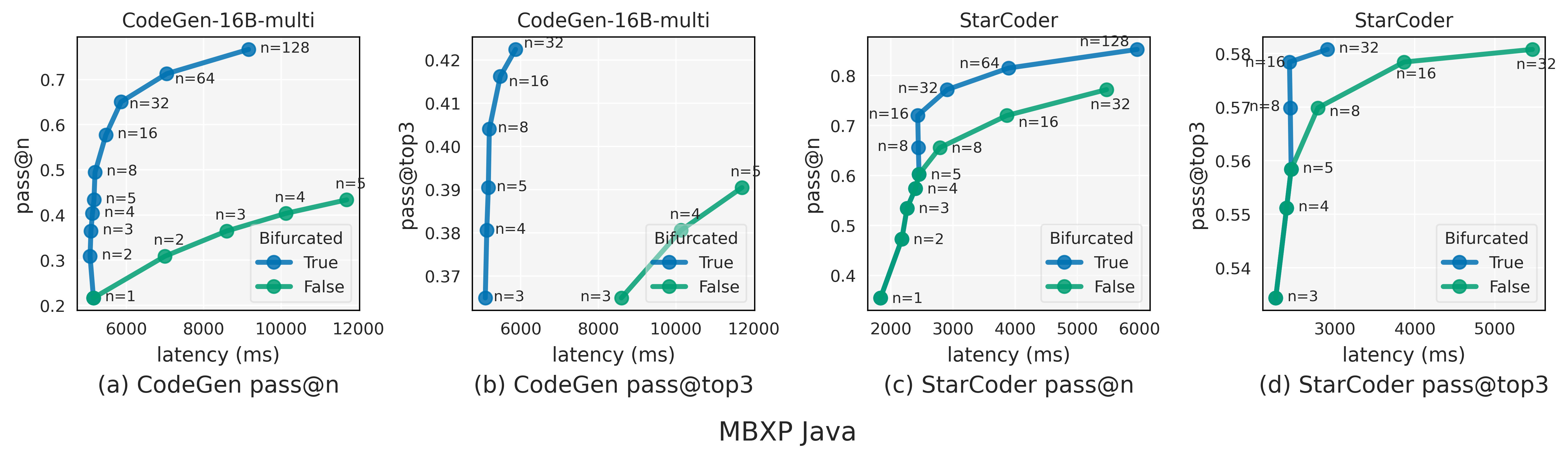}
    \caption{MBXP Java}
    \label{fig:application_java} 
\end{subfigure}

    \begin{subfigure}[t]{\plotwidth}
    \includegraphics[width=\textwidth]{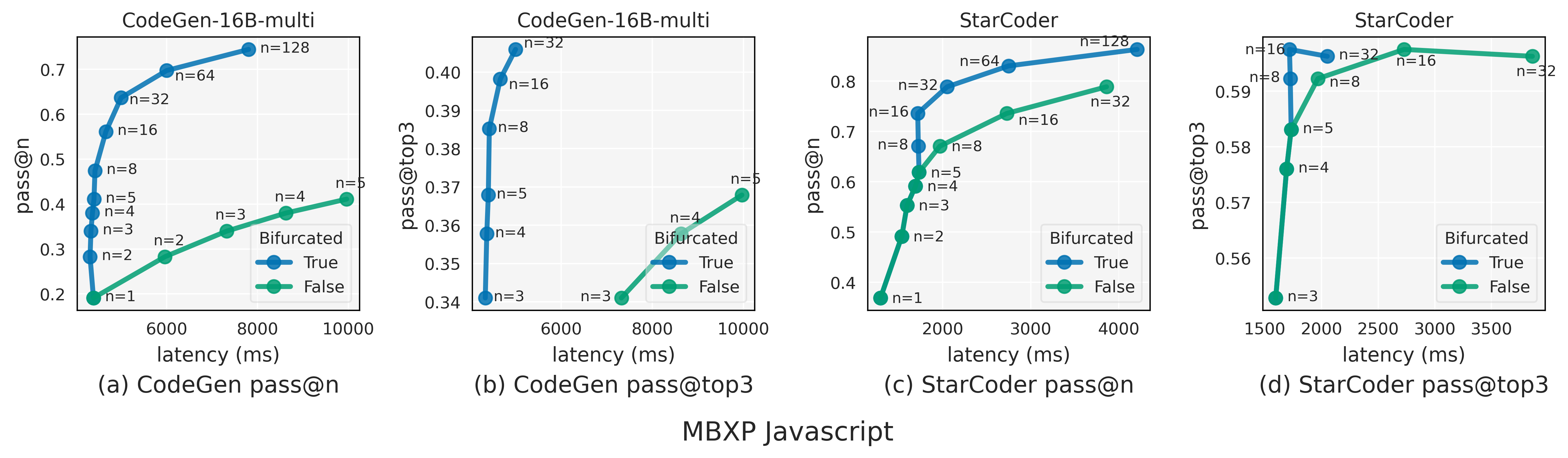}
    \caption{MBXP JavaScript}
    \label{fig:application_js} 
\end{subfigure}
    
    \caption{Bifurcated attention enables high batch sampling with minimal increase in latency with respect to batch size, resulting in more diverse and improved quality samples as indicated by pass@n and pass@top3 on MBXP-Java (Figure \ref{fig:application_java}). This trend extends to other evaluation such as JavaScript (Figure \ref{fig:application_js}) and Python (Figure \ref{fig:application_python}).
    }
    \label{fig:bifurcated_applications_extra}
\end{figure*}

\section{Applications: Additional Results}
We demonstrate additional results to the evaluation in Section \ref{sec:application} on MBXP-Java and MBXP-Javascript, in addition to the Python results. We replace CodeGen-16B-mono with CodeGen-16B-multi for the evaluation on Java and JavaScript and use the same StarCoder model. From Figure \ref{fig:bifurcated_applications_extra}, we observe similar trends as in Python (Figure \ref{fig:bifurcated_applications}), which furthers demonstrates the wide applicability of of bifurcated attention  in improving accuracy under latency-constrained scenarios.

\section{Compatibility with Speculative Decoding and Fast Decoding techniques}

Unlike standard auto-regressive decoding, fast decoding techniques such as  Speculative decoding\cite{speculative_decoding_deepmind,speculative_decoding_google}, Medusa 
\cite{medusa}, Lookahead \cite{lookahead}, and Eagle \cite{eagle} attempt to decode multiple tokens at each step. This reduces I/O bandwidth requirements because model parameters and KV cache are fetched only once per step and can be amortized across all generated tokens.

The fundamental principle behind these techniques is to first draft (or guess) a set of tokens (denoted as $n_g$) and then validate their accuracy by parallelly decoding with the model. After each step, up to a tokens (where $a \leq n_g$) may be accepted as valid, allowing for memory usage amortization across these accepted tokens. This approach is successful because decoding is primarily constrained by memory I/O.

The benefits of bifurcated attention are orthogonal to those of speculative sampling, leading to further memory I/O improvements. This can be observed by extrapolating per-step memory I/O costs from Section \ref{sec:memory_analysis} with $n_g$ raplacing $n$. Since $m >> n_g$ continues to hold, the advantages of bifurcated attention persist even when combined with speculative decoding.

\begin{table*}[ht]
    \small
    \centering
    \caption{Per-token generation latency (ms) with bifurcated attention compared to native Flash attention 2 kernel and Torch's SDPA attention kernel implementations, with and without using the torch compile option. Measurements are taken using a 7B parameter model (32 layers, 32 heads, hidden dimension = 4096) with multi-head attention. SDPA Math represents the default attention operations by Torch, while SDPA Flash utilizes Flash attention under the hood. "NC" refers to the use of non-contiguous memory allocation for the cache, allowing reuse of the cache from the prompt. Note that Flash attention kernels are currently not compatible with torch-compile. The experiment results below utilize an Nvidia H100 GPU.}
    \begin{tabular}{r|rrrrrr|rrr}
        \toprule
        ~ & \multicolumn{6}{c|}{without Torch Compile}     & \multicolumn{3}{c}{with Torch Compile} \\ 
        \midrule
        BS & \textbf{Bifurcated} & Flash2 & SDPA Math & SDPA Flash & Flash2  & SDPA Flash  & SDPA Math & \textbf{Bifurcated} & SDPA Math \\ 
        ~ & ~ & ~ & ~ & ~ & \multicolumn{1}{c}{(NC)} & \multicolumn{1}{c|}{(NC)} & \multicolumn{1}{c}{(NC)} & ~ & ~ \\ 
        \midrule
        & \multicolumn{9}{c}{Context Length : 8k}\\
        \midrule
        1 & 30.38 & 24.06 & 26.39 & 22.00 & 24.54 & 23.43 & 10.66 & 8.63 & 8.77 \\ 
        2 & 31.37 & 24.49 & 28.70 & 24.77 & 31.53 & 31.66 & 14.45 & 11.74 & 10.50 \\ 
        4 & 31.44 & 39.66 & 43.36 & 38.86 & 50.54 & 51.06 & 23.20 & 12.03 & 13.22 \\ 
        8 & 33.72 & 60.92 & 72.70 & 61.22 & 84.52 & 84.99 & 35.42 & 12.36 & 17.33 \\ 
        16 & 31.70 & 109.64 & 132.89 & 109.45 & 155.85 & 159.82 & 63.68 & 12.59 & 26.19 \\ 
        32 & 31.78 & 205.57 & 251.02 & 205.92 & 305.39 & 306.60 & 120.39 & 13.47 & - \\ 
        64 & 35.26 & OOM & OOM & - & 599.08 & 601.48 & 238.19 & 15.35 & - \\ 
        128 & 48.69 & - & - & - & 1183.46 & OOM & OOM & 19.56 & - \\ 
        256 & 75.21 & - & - & - & 1842.98 & - & - & 27.15 & - \\ 
        512 & 130.58 & - & - & - & - & - & - & 44.33 & - \\ 
        1024 & 242.73 & - & - & - & - & - & - & 81.14 & - \\ 
        2048 & 473.74 & - & - & - & - & - & - & OOM & - \\         \midrule
        & \multicolumn{9}{c}{Context Length : 16k}\\
        \midrule
        1 & 30.66 & 26.28 & 30.13 & 26.22 & 30.49 & 30.20 & 15.53 & 12.16 & 13.06 \\ 
        2 & 32.62 & 37.72 & 44.74 & 38.25 & 51.30 & 51.24 & 22.46 & 17.17 & 15.35 \\ 
        4 & 33.44 & 65.98 & 73.62 & 65.83 & 91.25 & 90.76 & 39.51 & 17.33 & 20.65 \\ 
        8 & 34.67 & 110.31 & 132.29 & 110.55 & 159.96 & 160.39 & 64.22 & 18.07 & 32.06 \\ 
        16 & 36.78 & 206.93 & 251.47 & 206.52 & 306.75 & 307.31 & 119.87 & 18.46 & OOM \\ 
        32 & 41.93 & OOM & OOM & OOM & 601.10 & 603.61 & 237.89 & 19.92 & - \\ 
        64 & 50.53 & - & - & - & 1195.35 & OOM & OOM & 22.96 & - \\ 
        128 & 68.31 & - & - & - & 1908.23 & - & - & 28.98 & - \\ 
        256 & 106.10 & - & - & - & OOM & - & - & 40.07 & - \\ 
        512 & 183.14 & - & - & - & - & - & - & 65.02 & - \\ 
        1024 & 339.74 & - & - & - & - & - & - & 117.75 & - \\ 
        2048 & 660.20 & - & - & - & - & - & - & OOM & - \\ 
        \midrule
        & \multicolumn{9}{c}{Context Length : 32k}\\
        \midrule
        1 & 39.97 & 37.67 & 44.94 & 37.46 & 67.44 & 67.30 & 30.39 & 20.90 & 19.80 \\ 
        2 & 48.61 & 55.94 & 69.22 & 55.86 & 156.61 & 156.35 & 47.63 & 29.34 & OOM \\ 
        4 & 49.77 & OOM & OOM & OOM & 300.47 & 300.97 & 90.19 & 29.73 & - \\ 
        8 & 51.31 & - & - & - & 567.93 & 568.81 & 152.19 & 30.30 & - \\ 
        16 & 54.92 & - & - & - & 670.21 & 672.42 & 290.59 & 30.66 & - \\ 
        32 & 62.28 & - & - & - & 1318.05 & 1323.25 & 569.74 & 32.15 & - \\ 
        64 & 75.22 & - & - & - & OOM & OOM & OOM & 35.25 & - \\ 
        128 & 101.18 & - & - & - & - & - & - & 41.44 & - \\ 
        256 & 159.09 & - & - & - & - & - & - & OOM & - \\ 
        512 & 277.05 & - & - & - & - & - & - & - & - \\ 
        1024 & OOM & - & - & - & - & - & - & - & - \\ 
        \bottomrule
    \end{tabular}
    \label{tab:mhagptfast}
\end{table*}

\section{Experiments with GPTFast}

The implementation of context-aware bifurcated attention in native PyTorch demonstrates significant reductions in parallel sampling latency for both multi-headed attention (MHA) and grouped query attention (GQA) architectures. Bifurcated attention, being context-aware and implemented natively in PyTorch, can directly benefit from PyTorch's compilation capabilities.

We observe Bifurcated attention outperforming FlashAttention2, especially for larger context lengths and higher degrees of tensor parallelism. Since Bifurcated attention is primarily targeting decode phase during inference, leveraging the efficiency of FlashAttention2 for the prefill (context encoding) step.

\subsection{In Comparison with FlashAttention}

\begin{table*}[t]
    \caption{Per-token generation latency (ms) with bifurcated attention compared to the native Flash attention kernel. Measurements are taken with a 7B parameter model (32 layers, 32 heads, hidden dimension = 4096, 8 kv heads) using grouped query attention. Note that Flash attention kernels are currently not compatible with torch-compile. In this table, "NC" refers to the use of non-contiguous memory allocation for the cache, allowing reuse of the cache from the prompt. The experiment results below utilize an Nvidia H100 GPU.}
\small
\label{tab:gqagptfast}
\centering
\begin{tabular}{r|rrr|rrr|rrr|rrr}
\toprule
BS   & \multicolumn{3}{c|}{\textbf{Bifurcated + Compile}} & \multicolumn{3}{c|}{\textbf{Bifurcated}} & \multicolumn{3}{c|}{Flash2} & \multicolumn{3}{c}{Flash 2 (NC)} \\
\midrule
 Context:     & \multicolumn{1}{c}{8k}           & \multicolumn{1}{c}{16k}         & \multicolumn{1}{c|}{32k}         &  \multicolumn{1}{c}{8k}           & \multicolumn{1}{c}{16k}         & \multicolumn{1}{c|}{32k}   &  \multicolumn{1}{c}{8k}           & \multicolumn{1}{c}{16k}         & \multicolumn{1}{c|}{32k}    &  \multicolumn{1}{c}{8k}           & \multicolumn{1}{c}{16k}         & \multicolumn{1}{c}{32k}       \\ \midrule
1    & 10.56        & 15.16       & 22.79       & 28.37     & 30.97    & 37.20   & 21.76   & 23.59   & 26.64  & 23.48     & 25.23    & 28.20     \\
2    & 11.35        & 15.99       & 23.72       & 29.53     & 32.16    & 37.47   & 22.46   & 23.78   & 26.82  & 39.93     & 28.53    & 45.70     \\
4    & 11.52        & 16.20       & 23.98       & 29.58     & 32.19    & 37.48   & 22.57   & 24.22   & 27.30  & 71.57     & 42.47    & 72.94     \\
8    & 11.79        & 16.61       & 24.59       & 29.58     & 32.41    & 38.12   & 22.65   & 24.03   & 28.36  & 126.35    & 70.01    & 127.96    \\
16   & 11.72        & 16.68       & 24.87       & 30.27     & 32.85    & 37.29   & 22.31   & 30.19   & OOM    & 240.96    & 130.77   & 245.81    \\
32   & 12.50        & 17.77       & 27.01       & 29.76     & 32.75    & 37.84   & 26.06   & OOM     &        & 468.93    & 244.54   & 467.61    \\
64   & 13.87        & 19.90       & 30.31       & 29.52     & 32.07    & 45.73   & OOM     &         &        & 403.08    & 482.71   & 463.55    \\
128  & 17.03        & 24.90       & 37.60       & 29.55     & 40.26    & 63.06   &         &         &        & 788.66    & 465.70   & 909.02    \\
256  & 24.38        & 33.76       & 52.06       & 40.07     & 59.42    & 96.28   &         &         &        &           & 915.89   & 1805.60   \\
512  & 39.08        & OOM         & OOM         & 65.74     & OOM      & OOM     &         &         &        &           & OOM      & OOM       \\
1024 & 72.24        &             &             & 118.57    &          &         &         &         &        &           &          &           \\
2048 & OOM          &             &             & 230.88    &          &         &         &         &        &           &          &          \\ \bottomrule
\end{tabular}
\end{table*}

    
    

FlashAttention is a highly efficient general-purpose fused attention kernel that is particularly effective during context encoding, as it avoids materializing the expensive-to-read-and-write $n \times n$ attention matrix in GPU memory. 

However, during incremental decoding with single-context batch sampling, native FlashAttention kernels are not as efficient because they are not designed to be context-aware. Specifically, if there are B batch indices of K,V cache that are duplicate in values due to the shared prefix, FlashAttention-2 (FA2) can use paged KV-Cache to refer and point them to the same KV-pairs for the prefix across a batch. Nevertheless, this does not prevent the FlashAttention kernel from performing multiple reads of the KV-pairs from the shared prefix.

Table \ref{tab:mhagptfast} shows that a context-aware approach such as bifurcated attention outperforms FlashAttention in parallel sampling scenarios, especially with an increasing number of parallel samples. Notably, the bifurcated attention kernel is utilized solely during the decode step, allowing the efficient FlashAttention2 kernel to be employed during the prefill step for context lengths up to 8192. Furthermore, while non-contiguous memory avoids out-of-memory issues during parallel sampling for non-context-aware kernels, bifurcated attention's memory setup, which maintains only one copy of the context and expands by reference across batch indices, achieves substantially lower latencies. However, the native FlashAttention2 implementation is not yet compatible with PyTorch's compilation capabilities.

In the future, it may be possible to combine bifurcated attention with FlashAttention to optimize the latency further.

\subsection{Trends with Grouped Querry Attention (GQA)}

For GQA architectures, bifurcated attention is able to help scale to very large inference workloads. Using PyTorch's compilation mode, the inference with bifurcated attention is much faster compared to FlashAttention2. Table \ref{tab:gqagptfast} presents the results for context lengths of 8K, 16K, and 32K tokens. Note that PyTorch's SDPA is not directly supported for GQA and thus not included in the comparison.

\subsection{Compatibility with Tensor Parallel (TP)}

Higher tensor parallelism is often required to handle higher inference workloads, as seen in Table~\ref{tab:tp2}. The proposed context-aware bifurcated attention method works out-of-the-box without additional modifications for tensor parallelism. With TP we get to work with much larger context lengths.

\begin{table*}[t]
\centering
\caption{Per-token generation latency (ms) Mistral 7B at different context lengths with TP=2 experimented on Nvidia's H100 GPU.}

\begin{tabular}{r|r|rrr}
\toprule
\multicolumn{1}{c|}{Context} & \multicolumn{1}{c|}{BS} & \multicolumn{1}{c}{SDPA} & \multicolumn{1}{c}{\textbf{Bifurcated}} & \multicolumn{1}{c}{Flash2} \\
\midrule
16384                                       & 16                                      & 131.46                           & 55.51                                  & 92.11                              \\
32640                                       & 8                                       & 133.85                           & 58.56                                  & 92.35                              \\
32640                                       & 16                                      & 246.53                           & 58.00                                  & 162.02                             \\
32640                                       & 32                                      & OOM                               & 57.86                                  & OOM                                 \\
32640                                       & 64                                      &                                   & 60.33                                  &                                     \\
32640                                       & 128                                     &                                   & 67.82                                  & \\
\bottomrule
\end{tabular}
\label{tab:tp2}
\end{table*}

\end{document}